\documentclass[a4paper,fleqn]{cas-sc}
\usepackage{apacite}
\usepackage[authoryear]{natbib}
\usepackage{hyperref}
\def\tsc#1{\csdef{#1}{\textsc{\lowercase{#1}}\xspace}}
\tsc{WGM}
\tsc{QE}
\tsc{EP}
\tsc{PMS}
\tsc{BEC}
\tsc{DE}

\begin{document}
\let\WriteBookmarks\relax
\def\floatpagepagefraction{1}
\def\textpagefraction{.001}

\shorttitle{3SHNet: Boosting Image-Sentence Retrieval via Visual Semantic-Spatial Self-Highlighting}

\shortauthors{Xuri Ge et~al.}

\title [mode = title]{3SHNet: Boosting Image-Sentence Retrieval via Visual Semantic-Spatial Self-Highlighting}                      



%
\author[1]{Xuri Ge}[type=editor,
                        auid=000,bioid=1,
                        prefix=,
                        role=,
                        orcid=0000-0002-3925-4951]



\ead{x.ge.2@research.gla.ac.uk}


\author[1]{Songpei Xu}[%
   role=, 
   suffix=,
   ]
\cormark[1]
\ead{s.xu.1@research.gla.ac.uk}

\author[2]{Fuhai Chen}[style=]  
\ead{chenfuhai3c@163.com}
\cormark[2]


\author[1]{Jie Wang}[%
   role=, 
   suffix=,
   ]
\ead{j.wang.9@research.gla.ac.uk}

\author[3]{Guoxin Wang}
\ead{guoxin.wang@zju.edu.cn}
\author[3]{Shan An}
\ead{anshan.tju@gmail.com}

\author[1]{Joemon M. Jose}[%
   role=, 
   suffix=,
   ]
   \cormark[2]
\ead{Joemon.Jose@glasgow.ac.uk}

\affiliation[1]{organization={University of Glasgow, School of Computing Science},
    addressline={Sir Alwyn Williams Building (SAWB)}, 
    city={Glasgow},
    postcode={G12 8RZ}, 
    country={United Kingdom}}
\affiliation[2]{organization={Fuzhou University, College of Computer and Data Science},
    city={Fuzhou},
    postcode={350116}, 
    state={Fujian},
    country={China}}

\affiliation[3]{organization={JD Health International Inc.},
    city={Beijing},
    postcode={100176}, 
    country={China}}

\cortext[cor1]{Equal contribution}
\cortext[cor2]{Corresponding author}



\begin{abstract}
In this paper, we propose a novel visual \textbf{S}emantic-\textbf{S}patial \textbf{S}elf-\textbf{H}ighlighting \textbf{Net}work (termed \textbf{\textit{3SHNet}}) for high-precision, high-efficiency and high-generalization image-sentence retrieval. 3SHNet highlights the salient identification of prominent objects and their spatial locations within the visual modality, thus allowing the integration of visual semantics-spatial interactions and maintaining independence between two modalities. This integration effectively combines object regions with the corresponding semantic and position layouts derived from segmentation to enhance the visual representation. And the modality-independence guarantees efficiency and generalization. Additionally, 3SHNet utilizes the structured contextual visual scene information from segmentation to conduct the local (region-based) or global (grid-based) guidance and achieve accurate hybrid-level retrieval. Extensive experiments conducted on MS-COCO and Flickr30K benchmarks substantiate the superior performances, inference efficiency and generalization of the proposed 3SHNet when juxtaposed with contemporary state-of-the-art methodologies. Specifically, on the larger MS-COCO 5K test set, we achieve 16.3\%, 24.8\%, and 18.3\% improvements in terms of rSum score, respectively, compared with the state-of-the-art methods using different image representations, while maintaining optimal retrieval efficiency. Moreover, our performance on cross-dataset generalization improves by 18.6\%. 
\end{abstract}


\maketitle 
\section{Introduction}
     Image-sentence retrieval, also known as cross-modal retrieval, serves the purpose of retrieving the most pertinent images or sentences based on a given query sentence or image. 
     This functionality holds immense significance in practical multimedia applications, such as image and text querying within recommendation systems \citep{yuan2023go}. Additionally, it plays a foundational role in enabling multi-modal retrieval within search engines \citep{he2011using}. The seamless integration of these modalities enhances the overall user experience and efficiency in information retrieval. Image-sentence retrieval is a critical task that poses significant challenges. One of the primary difficulties is the inherent semantic gap when measuring the precise semantic similarities between the visual and textual modalities. 
    Especially the textual semantics are more specific than the visual semantics since the sentence contains well-structured and explicitly-semantical instances, while the image involves rich visual semantic objects and a complex context. 
    The semantic gap is widening with the enormous development of the recent powerful language models, such as BERT \citep{devlin2018bert}.

    \begin{figure}[t] 
    	\centering
    	\includegraphics[width=0.7\linewidth]{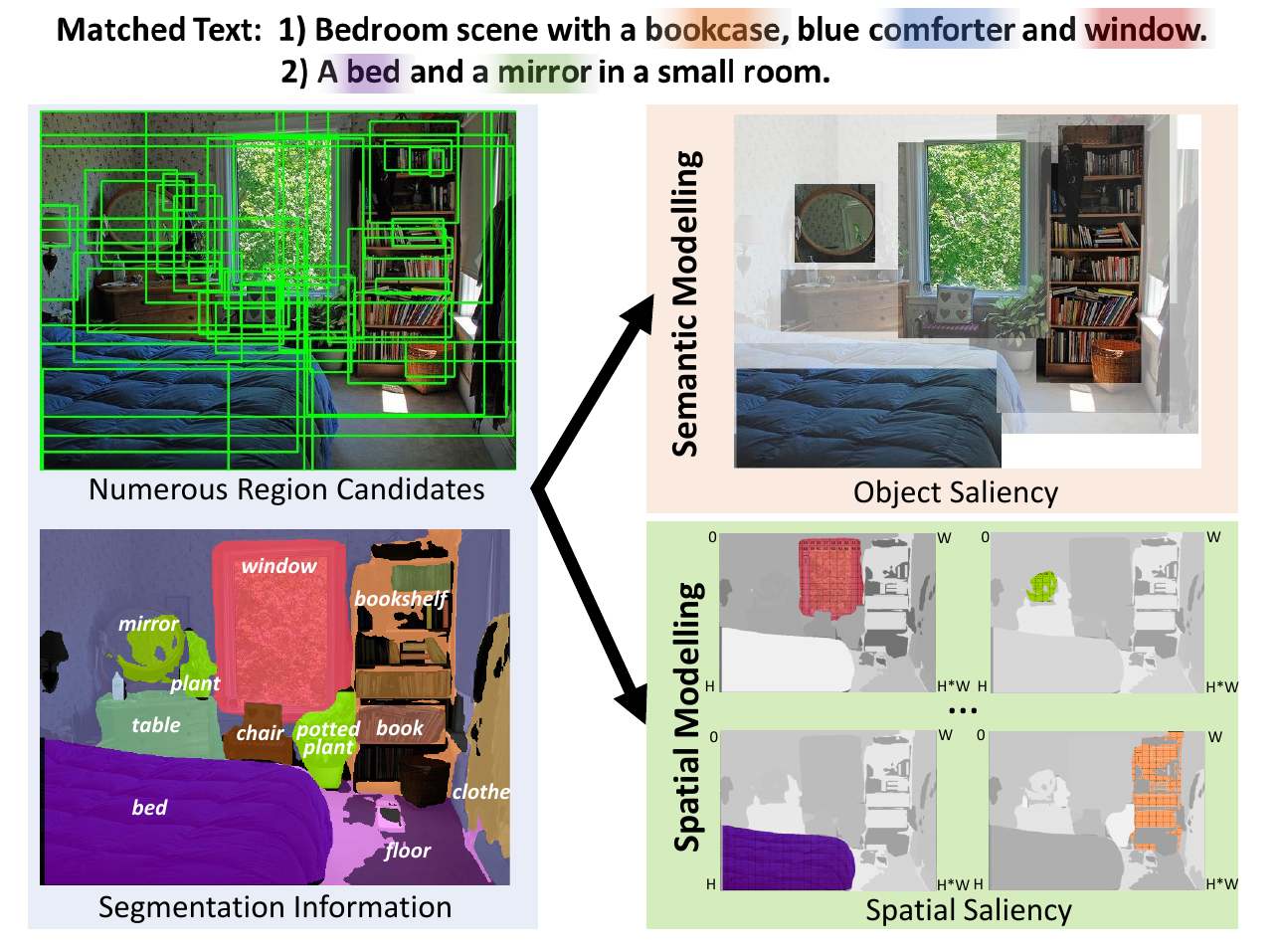}
        \caption{Segmentation is combined with the mass object regions to highlight the prominent objects and their locations. 
        }
    	\label{fig:motivation}
    \end{figure}

    Two popular schemes are developed to enhance the visual semantic features for image-sentence retrieval. 
    One is text-dependent visual representation learning \citep{chen2020imram,qu2021dynamic,ge2023cross,pan2023fine} while the other is hybrid-level visual representation enhancing \citep{zhang2020deep, zhang2022show, guo2023hgan}.
    On the one hand, text-dependent visual representation learning methods provide fine-grained correspondence across two modalities, where the highly relevant words are encoded into each object region by combination or attention. 
    For instance, in SCAN \citep{lee2018stacked}, text-aware visual features for image-to-sentence retrieval were crafted through an attention-based cross-modal interaction, which involved the model selectively attending to object regions associated with each word, amplifying the emphasis on visual semantics within the generated features.
    However, due to the joint embedding and deep interaction of visual and textual semantic features during both training and inference, these methods have two significant defects: (i) a massive redundant computation reduces the inference speed of retrieval since the visual feature of an image always needs to be recomputed once its similarity is estimated with a new sentence, and 
    (ii) due to the data differences and the differential effects of the different-dataset sentences on the visual representation, the trained model on one dataset cannot directly test well on another, which impedes cross-domain generalization.

    On the other hand, to gain a deeper understanding of images that includes global contextual information and fine-grained local representation, the latest studies \citep{zhang2020deep, chen2021learning, zhang2022show, wu2022improving} have proposed hybrid-level visual representations, which combine both local and global features. This approach allows for the complementary advantages of local fragments and global image features to be fully realized. These methods improve retrieval performance by capturing all possible semantics in global and local features without relying on textual guidance. For example, \citet{wu2022improving} fused local-level region and global-level features through a two-stage interaction strategy for image-text retrieval to extract a more holistic image representation. However, in human interaction, people tend to focus on prominent objects and their spatial locations \citep{walther2006interactions, madani2018human}, which is ignored in \citet{wu2022improving}.
    In image-sentence retrieval, the alignment between visual and textual data involves the consistencies of the prominent objects and their spatial locations, as the textual annotations inherently reflect the annotator's attention. Consequently, it is logical to incorporate human-like attention modelling into a visual representation, especially when the visual and textual modalities remain independent.

    Motivated by the aforementioned insights, the main research objectives of this study lie in two aspects: (i) to overcome the deep textual dependence for visual representation learning; and (ii) to explore the human-like attention from two aspects, \textit{i.e.} object semantic- and spatial-level, via a visual salient interaction schema in an end-to-end framework for cross-modal alignment. 
    To this end, we propose a visual semantic-spatial self-highlighting network (3SHNet) towards high-precision, high-efficiency, high-generalization image-sentence retrieval. 
    In particular, 3SHNet highlights the prominent objects and their spatial locations via the visual multi-modal interaction between the object regions and the segmentation results. 
    This approach emphasises the semantic and spatial saliencies arising from the intersection of the scattered regions and the structured visual scene information. It unifies them based on the correspondence between the semantic and position layouts derived from segmentation, as illustrated in Fig. \ref{fig:motivation}. 
     In fact, most of the current segmentation and multi-level object representation schemas are inspired by the human's visual attention perception, as revealed by \citet{vacher2023measuring,zhu2016beyond,anderson2018bottom}, where the segmentation schema is deemed to simulate the process of capturing the meaningful cues for human-like attention while the object representation schema simulates the process of formulating human-like attention. Thus, we argue that the fusion of segmentation and object-region features more comprehensively coincides with the information process of human-like attention since both of them are unique and indispensable parts of bio-inspired attention. 
    %
    Additionally, 3SHNet takes full advantage of the structured contextual visual scene information from segmentation to conduct the local (region-based) or global (grid-based) guidance to achieve accurate hybrid-level retrieval. 
    Thus, 3SHNet self-highlights the semantic-spatial saliencies to reduce the visual-textual gap while keeping the visual-textual modality independent. This allows 3SHNet to pre-extract and retain the visual features for each image before inference, even though there are new linguistic queries or new linguistic candidate sets and makes 3SHNet insensitive to sentence differences across different datasets.
    3SHNet can provide an efficient and effective paradigm to inspire existing real-world application scenarios, such as building a multi-modal recommendation system to help users quickly find the products or building a smart photo album to help users quickly find the images they need, etc.

    Outlined below are the contributions made by this paper: 
    \begin{itemize} 
     \item We explore the high-precision, high-efficiency, and high-generalisation image-sentence retrieval when the visual modality is independent of the textual modality. To achieve this,  human-like attention is forged within the visual modality. 
     
     \item We introduce a novel visual semantic-spatial self-highlighting network (3SHNet), where the segmentation is first utilized in image-sentence retrieval and interacted with the global and local visual features as the structured contextual guidance for semantic-spatial saliencies. 
     This allows for a unified interpretation of semantic-spatial saliencies over the segmentation maps. 
    \item  The proposed 3SHNet\footnote{Data and code are available at \href{https://github.com/XuriGe1995/3SHNet}{https://github.com/XuriGe1995/3SHNet.}} is verified to attain the state-of-the-art (SOTA) retrieval performances on two standard image-sentence retrieval benchmarks, \textit{i.e.}  MS-COCO and Flickr30K. Especially, 3SHNet is proven superior on the local-level, global-level, and hybrid-level visual features compared to the SOTAs, demonstrating its robustness. 
    Furthermore, the high-efficiency of 3SHNet is verified with higher inference speed compared to the SOTA, while the high-generalization is demonstrated on the cross-dataset training-testing setting. 
    
    \end{itemize}

\section{Related Work}
    Current studies to image-sentence retrieval can be categorized based on whether they incorporate visual-textual interaction or not: (i) cross-modal interaction retrieval \citep{li2021multi,qu2021dynamic,lee2018stacked,ge2023cross}, and (ii) modality-independent representation retrieval \citep{ma2023beat, liu2018dense,ge2021structured,chen2021learning,wang2018learning}. 
    Especially, most works explored the fine-grained correspondence of the cross-modality local-level representations, \textit{i.e.} employing cross-modal attention mechanisms to establish connections between image regions and sentence words. 
    For instance, DIME \citep{qu2021dynamic} adopted a multi-layer multiple cross-modality interaction framework by cross-modal attention-aware region/word aggregating and region-sentence/word-image correspondence learning. 
    Besides, more successes were also witnessed in recent large-scale pre-training visual-language models \citep{chen2020uniter,li2020oscar,li2021align}, which rely heavily on large-scale data-driven pattern and the powerful computation facilities, \textit{e.g.}, 400M image-text pairs and 256 V100 GPUs are used in \citet{radford2021learning}. 
    However, since they rely on the deep cross-modal interaction and need a new traversal computation cost once there is a new query of image or sentence, the retrieval takes a long inference time, which is hardly applied to real-life scenarios. 
    To address this issue, many recent modality-independent representation learning methods \citep{ge2021structured,chen2021learning, li2022multi} are proposed to encode visual and language information from both modalities into a joint embedding space without using any cross-attention interactions.

    However, due to the lack of textual guidance, most modality-independent representation retrieval methods \citep{chen2021learning,wu2022improving} proceed from the essence of modality-independent representation learning to elaborate visual-intrinsic feature enhancement models that can substitute textual guidance to reduce the semantic gap. 
   On the one hand, some methods \citep{cheng2022cross,long2022gradual} found the problem of missing relationships between visual objects and improved visual contextual representation by detecting object associations in scene graphs. 
    On the other hand, some latest works \citep{zhang2020deep,wu2022improving} started from various image representations to enhance the visual distinguishability via combining multiple levels of visual representation, \textit{i.e.,} local- and global-level image features.
    Indeed, they boost retrieval performance; however, they still neglect how humans pay attention to the prominent objects and their locations on the multiple levels of visual representation. 

    SAN \citep{ji2019saliency} introduced a global saliency detector to generate salience-weighted maps for images with additional supervision in a cross-modal interaction retrieval framework. 
    However, these saliency maps lack semantic and spatial information about objects and background information, and retrieval speed is still limited due to text dependency. 
    Visual semantic segmentation \citep{wu2022difnet,hu2023you,hu2023pseudo} can represent coarse-grained image semantics and precise spatial locations, which is usually used in many studies \citep{mousavian2015semantically,zhao2022learning}.  
    For example, DIFNet (\citep{wu2022difnet}) took segmentation feature as another visual information flow to improve image captioning performance, where the segmentation features are used independently of the original visual features.
    However, these methods fuse the segmentation information and weaken the most important characteristics of semantic segmentation results, such as accurate high-level category semantic information and its explicit spatial locations, which are also absent in salient object detection \citep{borji2015salient,ji2019saliency,pang2020multi}.
    
\begin{figure}[t] 
    	\centering
    	\includegraphics[width=1\linewidth]{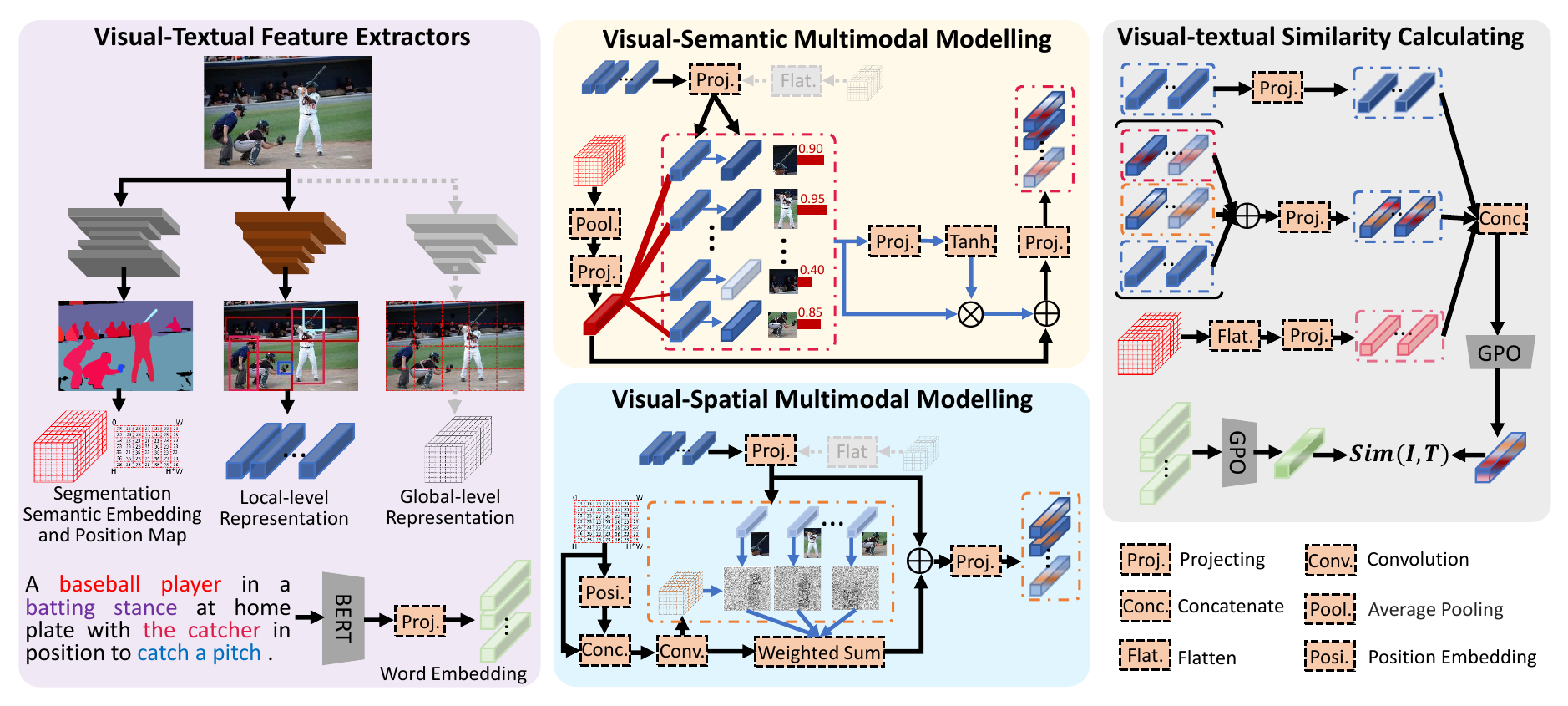}
     \caption{Illustration of the proposed 3SHNet. It mainly consists of visual-semantic modelling module (VSeM) and visual-spatial modelling module (VSpM), where the semantic feature and the position map of the segmentation are respectively imposed to guide the local- and global-level visual features in visual multimodal interactions.}
    	\label{fig:fig_overall}
    \end{figure}

\begin{table}[t]
\scriptsize
\begin{center}
\fontsize{9.5}{13}\selectfont
\renewcommand\tabcolsep{4.0pt}
\caption{Main notations and their definitions.} \label{tab:notations}
\begin{tabular}{l|l}
\hline

\hline
Notation & Definition \\ \hline 
$I$         &  an image in database        \\
$T$         &  a sentence in database        \\
$V^l$         &  the sub-region features of the image     \\
$K$         &  the number of detected sub-regions for each image \\
$V^g$         &  the grid-based image features of the image   \\
$V^s$       &  the semantic segmentation feature of the image  \\
$V^m$       &  the segmentation map for the image  \\
$E$         &  the word features for each sentence \\
$\alpha_i$ &  the salience weights of i-th sub-region in the image \\
$\{\dot{v}^l\}$    &  the attention-aware sub-region features \\
$\{\ddot{v}^l\}$    &  the fine-grained salience object representations after visual-semantic multimodal modelling \\

$PE()$         &  the position encoding function \\
$p$         & each pixel index in the image \\
$\ddot{V}^m$   & the refined feature of segmentation map \\
$\beta_{ij}$   & the position correspondence coefficient between the i-th region and j-th position index \\
$\ddot{V}^l$       & the visual-spatial representations after Visual-spatial multimodal modelling\\
\hline
         
\hline
\end{tabular} 
\end{center}
\end{table}     
\section{Approach}
   Fig. \ref{fig:fig_overall} illustrates the framework of our proposed 3SHNet. 3SHNet mainly consists of two innovative visual-semantic multimodal modelling (VSeM) and visual-spatial multimodal modelling (VSpM) modules to respectively highlight the prominent objects and their spatial locations via visual semantic-spatial salient interactions over segmentation maps. 
    They allow to self-highlight within the visual modality, effectively substituting attention cues from sentences, while maintaining high-efficiency and high-generalization.  
    For clarity, the main notations and their definitions throughout the paper are shown in Table \ref{tab:notations}.

    \subsection{Visual-Textual Feature Extractors} \label{MFX}
    
        For readability, we first introduce the feature extraction process of 3SHNet. 3SHNet use both the local- and global-level image representations to fully capture the comprehensive visual semantics.
        For fine-grained local-level image representation, we use bottom-up-attention network \citep{anderson2018bottom} to extract $K$ sub-region features $V^l$=$\{v^l_1, \cdots, v^l_K\}$ to cover the main semantic and represent the whole image $I$$\in$$\mathbb{R}^{H^I\times W^I \times 3}$. 
        For contextual global-level image representation, we use ResNeXt \citep{xie2017aggregated} to extract the grid-based features $V^g \in \mathbb{R}^{H\times W\times 2048}$ of the whole image after an AdaptiveAvgPool2d pooling operation \citep{lin2014network}. 
        To enhance visual representation of the prominent objects and their locations, we introduce the segmentation feature $V^s$$\in$$\mathbb{R}^{H\times W\times C^s}$ and segmentation map $V^m$$\in$$\mathbb{R}^{H^I\times W^I}$ for image $I$, where $H, W, H^I, W^I, C^s$ are respectively feature height, feature width, image height, image width and the semantic categories. These variables are extracted from an FPN-based network \citep{xiong2019upsnet} containing high-level object semantics and their corresponding spatial information.  
        
        For the sentences, we employ a pre-trained language model, specifically BERT \citep{devlin2018bert}, for the extraction of word-level textual representations, aligning with the contemporary approach in contemporary natural language processing research. 
        Specifically, we extract the textual features $E = \{e_1, e_2,...,e_N\}$ ($e_i \in$ $\mathbb{R}^{768}$) via the pre-trained BERT \citep{devlin2018bert} for each sentence, where $N$ is the number of words.
        To get a joint embedding space measured with visual representation, we utilize a fully connected (FC) layer \citep{liu2022mlp} to map the extracted word features into a D-dimensional space for each sentence $T$. 

    \subsection{Visual Semantic-Spatial Multimodal Modelling} \label{VSSM}
    
     Inspired by how humans observe and analyse images, we aim to reconstruct image representations from two perspectives: visual-semantic multimodal modelling and visual-spatial multimodal modelling. 
     Fig. \ref{fig:fig_overall} illustrates the training process of 3SHNet, where the image is transformed into either fine-grained local-level object region features or global-level grid-based features for semantic-spatial modelling. We use object region features as an example.

    \subsubsection{Visual-semantic multimodal modelling.} \label{vsem} 
        As shown in the middle of Fig. \ref{fig:fig_overall}, guided by the semantic segmentation features, the salience of the object region is highlighted. And they further interact together. 
        Specifically, the semantic segmentation feature $V^s$ is first projected into a coarse-grained D-dimensional semantic space by an FC layer after a global average pooling operation. 
       To distinguish the differentiation not only between center and marginal regions but also within marginal regions, we calculate the cosine similarities for all segmentation-region pairs and then obtain the salience weights $\{\alpha_i\}$ of all object regions guided by the segmentation features via a Sigmoid function \citep{mcculloch1943logical} (Softmax function \citep{chorowski2015attention} is also available as discussed in detail in Section \ref{sigvssoft}). 
        These can be formulated as:
        \begin{gather}
            \ddot{V}^s = {\rm FC (AvgPool}(V^s)),\  \ddot{V}^s \in \mathbb{R}^{D}, \\
                 \alpha_{i} = {\rm Sigmoid}(\frac{(\ddot{V}^s)^T (W^l v^l_i)}{\sqrt{D}||\ddot{V}^s||\ ||W^l v^l_i||}), i \in [1, K], \label{sigmoid}
            \end{gather}
        where $\alpha_{i}$ is the salience weight of i-th region and $W^{l}$ denotes the linear projection parameter shared for the i-th region feature mapping as key and value elements. After these, we can obtain the salience regions $\{\dot{v}^l_i\}$, where $\dot{v}^l_i=\alpha_{i} W^lv^l_i$.  Then we conditionally fuse the weighted fine-grained object features with the segmentation features to further enhance their semantic representation as follows:
        \begin{equation}
            \ddot{v}^l_i = \ddot{W}^l({\rm Tanh}(\dot{W}^l \dot{v}^l_i)\dot{v}^l_i + \ddot{V}^s),
        \end{equation}
        where $\dot{W}^l, \ddot{W}^l$ denote the linear projection parameters. 
        Finally, we obtain the fine-grained salience object representations $\ddot{V}^l=\{\ddot{v}^l_i\}$ combined with the semantically definitive segmentation features.
    

        \subsubsection{Visual-spatial multimodal modelling.} \label{vspm}
        Different from approaches \citep{li2022image,ge2023cross} that model spatial relationships among objects, we take advantage of explicit and salient object spatial segmentation boundaries in semantic segmentation maps to explore the positional relevance of visual local semantic and structured semantics. It also differs from Transformer \citep{vaswani2017attention}, which simply embeds the position information and attends among multiple components. 
        In particular, a positional encoding function \citep{vaswani2017attention} based on a trigonometric function is applied to embed each pixel index $p\in [1, H^I\times W^I]$ of the segmentation map $V^m\in \mathbb{R}^{H^I\times W^I}$ for an image $I$ in a dense vector, as follows:
        \begin{equation}
        \begin{aligned}
        PE_j(p)=\left\{
            \begin{aligned}
            &{\rm Sin}(p/10000^{j/d}), \  &if\ j\ is\ even, \\
            &{\rm Cos}(p/10000^{j/d}), \ &if\ j\ is\ odd,   
            \end{aligned}
            \right.
        \end{aligned}
        \end{equation}
        where $j\in[1,d]$ and $d$ is dimensions of the positional embedding. 
        As shown in the middle of Fig. \ref{fig:fig_overall}, we concatenate the positional embedding and the normalized segmentation map into a new dense vector $\dot{V}^m \in \mathbb{R}^{H^I\times W^I \times (d+1)}$. Then a convolutional layer is used to down-sample the vector, which can further refine the positional embedding with certain semantics. 
        The above process can be formulated as:
        \begin{gather}
            \dot{V}^m = {\rm Concat}({\rm PE}(V^m_{p}), V^m), \\
            \ddot{V}^m = {\rm Conv2d}(\dot{V}^m),
        \end{gather}
        where $\ddot{V}^m $$\in$$ \mathbb{R}^{H^p\times W^p \times C^p}$ ($C^p $$\ll$$ C^s$ to keep model efficient) serve as the key and value for next visual-spatial attention modelling.
        Similarly, each region feature $v_i^l$ is projected into the same dimension with $\ddot{V}^m$ as a query and then calculates the position correspondence coefficient $\beta_i$ with the refined positional embedding $\ddot{V}^m$ by a per-dimension $\lambda$-smoothed Softmax \citep{chorowski2015attention}. A spatially concentrated feature will be obtained for each object, as follows:
         \begin{gather}
            c_{ij} = \frac{(U^lv_i^l)(\ddot{V}^m_j)^T}{||U^lv_i^l||\ ||\ddot{V}^m_j||}, i \in [1,K], j \in [1,H^p\times W^p],\\
            \beta_{ij} = \frac{exp(\lambda c_{ij})}{\sum_{j=1}^{H^p\times W^p} exp(\lambda c_{ij})},   \\
            \dddot{v}_i^m = \sum\nolimits_{j=1}^{H^p\times W^p} \beta_{ij} \ddot{v}^m_j, 
        \end{gather}
        where $U^l$ denotes the linear projection parameter. Finally, we combine the spatial embeddings and the corresponding region features with a mapping parameter $\dddot{U}^l$ as visual-spatial representations $\dddot{V}^l=\{\dddot{v}^l_i\}$. 
        \vspace{-0.2em}
        \begin{gather}
            \dddot{v}_i^{l} = \dddot{U}^l(\dddot{v}_i^m+ U^lv_i^l)
        \end{gather}

      Note that we focus on estimating the position relevance of the high-level local object semantics and the segmentation semantics to acquire the local positional representation. Additionally, the local positional representation is associated with local object semantic features in Eq.(10) to enhance the semantic-spatial representation. All of these are different from the cross-attention in Transformer. 

    \subsection{Feature Aggregation and Objective Function} \label{OBJ}
        To calculate the visual-textual similarities, as shown in the right of Fig. \ref{fig:fig_overall}, we aggregate multiple representations into a measurably uniform embedding space. 
        For the visual aggregation, we first combine the semantically enhanced representations, the spatially enhanced representations and the original region features and project them as the visual semantic-spatial representations. 
        Then, we aggregate these fine-grained local-level visual features, visual semantic-spatial features, and semantic segmentation features for each image by a popular generalized pooling operator (GPO) \citep{chen2021learning}, since GPO automatically seeks the best pooling function compared to the traditional pooling strategy, \textit{e.g.} max-pooling.  Similarly, we use GPO to obtain the final textual embedding space for textual aggregation. 
        Finally, the similarity scores are calculated between two-modality representations.
        During training, a bidirectional triplet ranking loss with hard negative mining \citep{faghri2017vse++} is adopted as the optimization objective, 
        as follows:
        \begin{equation}
            \begin{aligned}
                \mathcal{L}_{rank} = \sum_{(I,T)} \{ {\rm max}[0, \gamma - {\rm Cos}(I,T)+ {\rm Cos}(I,\bar{T})] \\  
                + {\rm max}[0, \gamma - {\rm Cos}(I,T)+ {\rm Cos}(\bar{I},T)]\} 
            \end{aligned}
        \end{equation}    
        where $\gamma$ is a margin constraint.

\section{Experiments}
    \subsection{Experiment Setup}
        \subsubsection{Dataset.}
        To validate the effectiveness of our approach, we undertake comprehensive experimentation on two widely recognized datasets, \textit{i.e.,} MS-COCO \citep{lin2014microsoft} and Flickr30k \citep{young2014f30k}.  
        Flickr30K \citep{young2014f30k} comprises more than 31,000 images, including 29, 000 training images, 1,014 validation images, and 1,000 test images. 
        Moreover, MS-COCO \citep{lin2014microsoft} consists of over 123,000 images with 82,738 allocated for training, 5,000 for testing, and an additional 5,000 for validation purposes. 
        In both benchmarks, five sentences for each image are supplied, each originating from a different AMT worker.
        
        \subsubsection{Evaluation metrics.}
        Following the mainstream \citep{faghri2017vse++,ge2023cross,chen2021learning,li2023towards}, we evaluate the numerical efficacy of all approaches by the widely employed recall metrics, Recall@Q (Q=1, 5, 10), indicating the percentage of ground-truth instances successfully matched among the top Q rankings. 
        Furthermore, we follow the standard rSum metric to calculate the summation of all six recall rates, thereby substantiating the comprehensive performance assessment. 
        \begin{equation}
          rSum = \underbrace{(Recall@1 + Recall@5 + Recall@10)}_{(Image-to-Sentence)} + \underbrace{(Recall@1 + Recall@5 + Recall@10)}_{(Sentence-to-Image)},
        \end{equation}
        In addition, we apply commonly used Kpps \citep{wang2022coder} to evaluate the model inference speed, which means the number of image/sentence queries completed per second.
        %

    \subsubsection{Implementation details.} \label{ID}
        The whole network except the offline visual extractors is implemented with PyTorch on a single TITAN RTX GPU using AdamW optimizer with weight decay factor 10e-4, where the learning rate is set to 5e-4 initially. 
        The maximum epoch number is configured at 25, accompanied by a mini-batch size of 256. 
        The joint embedding space possesses a dimensionality of 1024.
        The margin parameter $\gamma$ is specified as 0.2.
        We used the same pre-extracted features as the compared methods to guarantee the fairness. 
        Specifically,  we used the pre-extracted local-level region features, global-level grid features and segmentation results from Faster-RCNN \citep{ren2015faster}, ResNext-101 \citep{xie2017aggregated}, and UPSNet \citep{xiong2019upsnet}, respectively. The single-thread feature extraction speeds of these visual encoders are 2.1 FPS for Faster-RCNN, 2.1 FPS for ResNext-101 and 10.5 FPS for UPSNet, respectively. 
        These models are pre-trained on small-scale datasets, such as ImageNet \citep{russakovsky2015imagenet} and Visual Genome \citep{krishna2017visual}, \textit{etc}. 
        The main fine-grained pre-extracted local- and global-level visual features are the same as the compared methods to guarantee fairness. 
        For the local-level visual feature, the strategy entails choosing 36 regions (K=36) characterized by the highest confidence scores in object detection \citep{ren2015faster}, containing some redundant and useless regions. 
        For the global-level visual feature, the size of grid features is 7$\times$7$\times$2048.
        Following \citep{wu2022difnet}, the semantic segmentation feature size is 7$\times$7$\times$133, in which the dimension of 133 is logit corresponding to object categories. 
        The size of the segmentation map is resized to 64$\times$64 to reduce calculations.
        Note that although we use the extra segmentation features, the off-line speed testing is not influenced since it excludes the feature extraction process in a practical way. 
        The segmentation features without our deep feature interactions have inapparent gains on the retrieval performance according to the comparisons in later ablation studies, and Tab. \ref{tab:tab1_coco_1K} and Tab. \ref{tab:tab1_coco_5K} also show the comparisons between ours and the embeddable SOTA method (VSE$\rm \infty^{\tiny\text{w/ \ Seg.}}$ in same mini-batch) with such extra features.

\begin{table}[t]
\begin{center}
\fontsize{8}{10}\selectfont
\renewcommand\tabcolsep{5pt}
\caption{Comparisons of performances on MS-COCO 1K (5-folds) test set. The best results are highlighted in bold typeface. $*$ indicates the performance metrics attributed to the ensemble model.  For clearer comparison, the ensemble model is shown with a blue background and the improvements of the best contrasting method with underline are marked.} \label{tab:tab1_coco_1K}
\begin{tabular}{c|c|cccccc|c}
\hline

\hline
\multicolumn{1}{c|}{\multirow{2}{*}{Type}} &  \multicolumn{1}{c|}{\multirow{2}{*}{Method}}    & \multicolumn{3}{c}{Image-to-Sentence} & \multicolumn{3}{c|}{Sentence-to-Image}      & \multirow{2}{*}{rSum}  \\
& \multicolumn{1}{c|}{}                        & Recall@1        & Recall@5       & Recall@10      & Recall@1  & Recall@5 & \multicolumn{1}{c|}{Recall@10} &               \\ \hline

&IMRAM$^*$\citep{chen2020imram}   & {76.7}    & {95.6}    & {98.5}     &   61.7    &   89.1   &  {95.0}  & 516.6     \\ 
& VSE$\rm \infty$ \citep{chen2021learning} & 79.7    &96.4    & 98.9     &  64.8     &  91.4     &  96.3    &  527.5  \\
& DIME$^*$ \citep{qu2021dynamic}  & 78.8     & 96.3    & 98.7     & 64.8     & 91.5     & 96.5     & 526.6  \\  
&  VSRN++$^*$  \citep{li2022image} & 77.9    & 96.0     & 98.5   &  64.1     & 91.0   & 96.1  & 523.6\\ 
& NAAF$^*$  \citep{zhang2022negative}  & {80.5}    & 96.5     & 98.8     & 64.1     & 90.7     & {96.5}      & 527.2  \\ 
&  AME$^*$  \citep{li2022action}   & 79.4   & {96.7}    & {98.9}     & {65.4}     & 91.2     & 96.1      & {527.7}  \\ 
&  CMSEI$^*$  \citep{ge2023cross} & {81.4}   & {96.6}    & {98.8}     & {65.8}     & {91.8}     & {96.8}      & {531.1}  \\ %
& CHAN  \citep{pan2023fine} & {81.4}    & {96.9}    & {98.9}     & {66.5}     & {92.1}     & 96.7     & {532.6}   \\ 
& RCTRN* \citep{li2023reservoir}   & 79.4 & 96.6 & 98.3  & 66.9 & 92.2 & 96.8 & 530.2 \\
&  KIDRR* \citep{xie2023unifying} & 80.9 & 96.5 & 99.0 & 65.0 & 91.1 & 96.1 & 528.6 \\
& DCIN* \citep{li2023towards}  &  {81.4} & 96.8 & 99.0 & 66.1 & 92.1 & 96.6 & 532.0 \\
&  \underline{EKDM* \citep{yang2023external}} &  {81.4} & 96.7 & \textbf{99.4} & {68.5} & \textbf{93.5} & \textbf{97.6} & {537.1} \\
& DCIN* \citep{li2023towards}  &  {81.4} & 96.8 & 99.0 & 66.1 & 92.1 & 96.6 & 532.0 \\
&  {MKTLON* \citep{qin2024multi}} &  81.8 & 96.6 & 98.8 & 66.1 & 91.6 & 96.6 & 531.5 \\

\cline{2-9} 
 & \multicolumn{1}{c|}{ \textbf{3SHNet}}     &   {83.1} &   {97.2} & 
   {99.3} &   {68.7} &  {92.4} &  96.6 &   {537.3}  \\
\multirow{-14}{*}{\rotatebox{90}{Region}} & \multicolumn{1}{c|}{\cellcolor{blue!10}\textbf{3SHNet$^*$}} & \cellcolor{blue!10}\textbf{84.3$_{\tiny\text{+2.9}}$} & \cellcolor{blue!10}\textbf{97.3$_{\tiny\text{+0.6}}$} & \cellcolor{blue!10}{99.1$_{\tiny\text{-0.3}}$} & \cellcolor{blue!10}\textbf{69.7$_{\tiny\text{+1.2}}$} & \cellcolor{blue!10} {93.1$_{\tiny\text{-0.4}}$} & \cellcolor{blue!10} {97.0$_{\tiny\text{-0.6}}$} & \cellcolor{blue!10}\textbf{540.5$_{\tiny\text{+3.4}}$}   \\ 

\hline

& SCO \citep{huang2018learning}   & 69.9   & 92.9   & 97.5   & 56.7   & 87.5   & 94.8   & 499.3 \\ 
&  \underline{ VSE$\rm \infty$  \citep{chen2021learning}}  & 80.4    & 96.8    & 99.1     & 66.4     & 91.1     & 95.5     & 531.6 \\
\cline{2-9} 
 & \multicolumn{1}{c|}{\textbf{3SHNet}}     &  {83.1} &  {97.5} &  {99.1} &  {69.8} &  {92.7} &  {96.8} &  {538.9} \\
\multirow{-4}{*}{\rotatebox{90}{Grid}} &  \multicolumn{1}{c|}{\cellcolor{blue!10}\textbf{3SHNet$^*$}}  & \cellcolor{blue!10}\textbf{84.5$_{\tiny\text{+4.1}}$} & \cellcolor{blue!10}\textbf{97.7$_{\tiny\text{+0.9}}$} & \cellcolor{blue!10}\textbf{99.3$_{\tiny\text{+0.2}}$} & \cellcolor{blue!10}\textbf{70.6$_{\tiny\text{+4.2}}$} & \cellcolor{blue!10}\textbf{93.3$_{\tiny\text{+2.2}}$} & \cellcolor{blue!10}\textbf{97.1$_{\tiny\text{+1.0}}$} & \cellcolor{blue!10}\textbf{542.5$_{\tiny\text{+10.9}}$}  \\ 

\hline
& CRGN \citep{zhang2020deep} & 73.8  & 95.6  & 98.5  & 60.1  & 88.9  & 94.5  & 511.4 \\
& MLSL \citep{li2021multi} & 77.1 & 96.3 & 98.6 & 63.8 & 90.1 & 95.9  & 521.8 \\
&   VSE$\rm \infty$ \citep{chen2021learning}  &  82.2  &  97.5  &  \textbf{99.5}  &  68.1  &   {92.9}  &  97.2  &  537.4   \\
& CMCAN \citep{zhang2022show}  & 81.2  & 96.8  & 98.7  & 65.4  & 91.0  & 96.2  & 529.3 \\ 
&  \underline{ Imp.$\rm ^*$   \citep{wu2022improving}}     &   {83.7} &   {97.7} &  99.1 &   {68.4} &  92.8 &  \textbf{97.5} &   {539.2} \\
&  RAAN* \citep{wang2023rare} &  76.8 &  96.4 &  98.3 &  61.8 &  89.5 &  95.8 &  518.6 \\
&  HGAN \citep{guo2023hgan}   & 81.1 & 96.9 & 99.0 & 67.4 & 92.2 & 96.6 & 533.2  \\ \cline{2-9} 
&  VSE$\rm \infty^{\tiny\text{w/ \ Seg.}}$   & 83.1  & 97.6  & {99.5}  & 68.9  & {93.0}  & 97.2  & 539.3   \\
\cline{2-9} 
 & \multicolumn{1}{c|}{ \textbf{3SHNet}}    &  {85.0} & 
  \textbf{97.7} & 
   {99.2} &  {71.2} &  {93.5} & 
  {97.2} &  {543.7}\\
 \multirow{-10}{*}{\rotatebox{90}{Region+Grid}}  &  \multicolumn{1}{c|}{\cellcolor{blue!10}\textbf{3SHNet$^*$}} & \cellcolor{blue!10}\textbf{85.8$_{\tiny\text{+2.1}}$} & \cellcolor{blue!10}\textbf{97.7$_{\tiny\text{+0.0}}$} & \cellcolor{blue!10} {99.3}$_{\tiny\text{+0.2}}$ & \cellcolor{blue!10}\textbf{71.8$_{\tiny\text{+3.4}}$} & \cellcolor{blue!10}\textbf{93.7$_{\tiny\text{+0.9}}$} & 
 \cellcolor{blue!10}{ {97.4}$_{\tiny\text{-0.1}}$} &
 \cellcolor{blue!10}\textbf{545.7$_{\tiny\text{+6.5}}$}  \\ 

\hline

\hline
\end{tabular}
\end{center}
\end{table}
\begin{table}[t]
\begin{center}
\fontsize{8}{10}\selectfont
\renewcommand\tabcolsep{5pt}
\caption{Comparisons of performances on larger 5K test set.  The best results are highlighted in bold typeface.  $*$ indicates the performance of the ensemble model. For clearer comparison, the ensemble model is shown with a blue background and the improvement of the best contrasting method (with underline) is marked.} \label{tab:tab1_coco_5K}
\begin{tabular}{c|c|cccccc|c}
\hline

\hline
\multicolumn{1}{c|}{\multirow{2}{*}{Type}} &  \multicolumn{1}{c|}{\multirow{2}{*}{Method}}    & \multicolumn{3}{c}{Image-to-Sentence} & \multicolumn{3}{c|}{Sentence-to-Image}      & \multirow{2}{*}{rSum} \\
& \multicolumn{1}{c|}{}                      & Recall@1        & Recall@5       & Recall@10      & Recall@1  & Recall@5 & \multicolumn{1}{c|}{Recall@10} &                      \\ \hline
&IMRAM$\rm ^*$  \citep{chen2020imram}    &  53.7    & 83.2    & 91.0    & 39.7    & 69.1    & 79.8    & 416.5   \\ 
& VSE$\rm \infty$ \citep{chen2021learning}  &  58.3  &  85.3  &   92.3  &  42.4  &  72.7  &   83.2    &   434.3  \\
& DIME$^*$ \citep{qu2021dynamic}   & 59.3    & 85.4    & 91.9    & 43.1    & 73.0    & 83.1    & 435.8 \\  
&  VSRN++$^*$  \citep{li2022image} & 54.7  &  82.9   & 90.9   & 42.0   & 72.2   & 82.7   & 425.4  \\  
& NAAF$^*$  \citep{zhang2022negative}  & 58.9    &  85.2    &  92.0    &  42.5    &  70.9    &  81.4    &  430.9  \\ 
&  AME$^*$  \citep{li2022action}   & 59.9    & 85.2    & 92.3    &  {43.6}    & 72.6    & 82.7    & 436.3 \\ 
&   CMSEI$^*$  \citep{ge2023cross}    &     {61.5}   &   {86.3}   &   {92.7}  &    {44.0}  &   {73.4}   &   {83.4}   &   {441.2}\\ %
& \underline{ CHAN  \citep{pan2023fine} } &  59.8  &   {87.2}  &    {93.3}  &   {44.9}  &   {74.5}  &    {84.2}    &    {443.9}  \\

& RCTRN* \citep{li2023reservoir}  &  57.1 &  83.4 &  91.9 &  43.6 &  71.9 &  83.7  &  431.6 \\
&  KIDRR* \citep{xie2023unifying}  & 60.3 & 86.1 & 92.5 & 43.5 & 72.8 & 82.8 & 438.0 \\
& DCIN* \citep{li2023towards} & 60.8  & 86.3  & 93.0  & 44.0  & 74.6  & 84.3  & 443.0 \\
&  {MKTLON* \citep{qin2024multi}} &  61.4 & 86.7 & 92.8 & 44.3 & 73.9 & 83.7 & 442.8 \\
\cline{2-9} 
 & \multicolumn{1}{c|}{ \textbf{3SHNet}}     &   {63.8}  &   {88.1}  &   {94.0}  &   {47.0}  &   {76.6}  &   {85.4}  &   {454.9} \\
\multirow{-13}{*}{\rotatebox{90}{Region}} & \multicolumn{1}{c|}{\cellcolor{blue!10}\textbf{3SHNet$^*$}} & \cellcolor{blue!10}\textbf{65.3{$_{\tiny\text{+5.5}}$}}  & \cellcolor{blue!10}\textbf{88.8{$_{\tiny\text{+1.6}}$}}  & \cellcolor{blue!10}\textbf{94.1{$_{\tiny\text{+0.8}}$}}  & \cellcolor{blue!10}\textbf{48.2{$_{\tiny\text{+3.3}}$}}  & \cellcolor{blue!10}\textbf{77.5{$_{\tiny\text{+3.0}}$}}  & \cellcolor{blue!10}\textbf{86.3{$_{\tiny\text{+2.1}}$}}  & \cellcolor{blue!10}\textbf{460.2{$_{\tiny\text{+16.3}}$}}  \\ 

\hline

& SCO \citep{huang2018learning}    & 42.8   & 72.3   & 83.0   & 33.1   & 62.9   & 75.5   & 369.6\\ 
&   \underline{VSE$\rm \infty$  \citep{chen2021learning} }  &  59.1   &  85.9   &  92.8   &  44.1   &  74.1   &  84.0   &  440.0\\
\cline{2-9} 
 & \multicolumn{1}{c|}{ \textbf{3SHNet}}     &   {64.1} &   {88.9} &   {94.3} &   {48.0} &   {77.4} &   {86.3} &   {459.0}\\
\multirow{-4}{*}{\rotatebox{90}{Grid}} &  \multicolumn{1}{c|}{\cellcolor{blue!10}\textbf{3SHNet$^*$}}   & \cellcolor{blue!10}\textbf{66.2{$_{\tiny\text{+7.1}}$}}  & \cellcolor{blue!10}\textbf{89.8{$_{\tiny\text{+3.9}}$}}  & \cellcolor{blue!10}\textbf{94.7{$_{\tiny\text{+1.9}}$}}  & \cellcolor{blue!10}\textbf{49.0{$_{\tiny\text{+4.9}}$}}  & \cellcolor{blue!10}\textbf{78.3{$_{\tiny\text{+4.2}}$}}  & \cellcolor{blue!10}\textbf{86.8{$_{\tiny\text{+2.8}}$}}  & \cellcolor{blue!10}\textbf{464.8{$_{\tiny\text{+24.8}}$}}  \\ 

\hline
& CRGN \citep{zhang2020deep}  & 51.2  & 80.6  & 89.7  & 37.4 & 68.0  & 79.5  & 406.4\\
&   VSE$\rm \infty$ \citep{chen2021learning}  &  62.5   &  87.8   &   {94.0}   &  46.0   &  75.8   &   {85.7}   &  451.8 \\
& CMCAN \citep{zhang2022show}    & 61.5   & - & 92.9   & 44.0   & -  & 82.6   & -\\
&    \underline{ Imp.$\rm ^*$   \citep{wu2022improving} }    &   {63.5}  &    {87.9}  &   93.5  &    {46.8}  &   {76.1}   &  85.1   &   {452.9}  \\ 
&  HGAN \citep{guo2023hgan}   & 60.0  & 85.8  & 92.8  & 45.4  & 75.3  & 85.1  & 444.4 \\ \cline{2-9} 
&  VSE$\rm \infty^{\tiny\text{w/ \ Seg.}}$   & 63.9   & 88.3   & {94.2}   & 47.8   & 76.9   & {86.0}   & 457.1 \\
\cline{2-9} 
 & \multicolumn{1}{c|}{ \textbf{3SHNet}}   &   {67.1} &   {89.8} &   {95.2} &   {49.9} &   {78.8} &   {87.2} &   {468.0}\\
 \multirow{-8}{*}{\rotatebox{90}{Region+Grid}} &  \multicolumn{1}{c|}{\cellcolor{blue!10}\textbf{3SHNet$^*$}}   & \cellcolor{blue!10}\textbf{67.9{$_{\tiny\text{+4.4}}$}}  & \cellcolor{blue!10}\textbf{90.5{$_{\tiny\text{+2.6}}$}}  & \cellcolor{blue!10}\textbf{95.4{$_{\tiny\text{+1.9}}$}}  & \cellcolor{blue!10}\textbf{50.3{$_{\tiny\text{+3.5}}$}}  & \cellcolor{blue!10}\textbf{79.3{$_{\tiny\text{+3.2}}$}}  & \cellcolor{blue!10}\textbf{87.7{$_{\tiny\text{+2.6}}$}}  & \cellcolor{blue!10}\textbf{471.2{$_{\tiny\text{+18.3}}$}}  \\ 

\hline

\hline
\end{tabular}
\end{center}
\end{table}      
    \subsection{Quantitative Comparison}
       We report the performances of 3SHNet on MS-COCO and Flick30K with the local-level region-based image features, the global-level grid-based image features and the hybrid-level (region+grid) image features, respectively in Tab. \ref{tab:tab1_coco_1K}, Tab. \ref{tab:tab1_coco_5K} and Tab. \ref{tab:tab1_f30k}, compared with the corresponding state-of-the-art studies, including (1) region-based methods, \textit{i.e.,}  IMRAM* \citep{chen2020imram}, SGRAF \citep{diao2021similarity}, VSE$\rm \infty$ \citep{chen2021learning}, DIME* \citep{qu2021dynamic}, NAAF* \citep{zhang2022negative}, CHAN \citep{pan2023fine}, CMSEI* \citep{ge2023cross}, DCIN* \citep{li2023towards}, RCTRN* \citep{li2023reservoir}, KIDRR* \citep{xie2023unifying} and MKTLON* \citep{qin2024multi} \textit{etc.,} (2) grid-based methods, \textit{i.e.,} SCO* \citep{huang2018learning} and VSE$\rm \infty$ \citep{chen2021learning}, and (3) region-grid-based methods, \textit{i.e.,} CRGN \citep{zhang2020deep}, MLSL \citep{li2021multi}, CMCAN \citep{zhang2022show}, Imp.* \citep{wu2022improving}, RAAN* \citep{wang2023rare} and HGAN* \citep{guo2023hgan}. 
       Our 3SHNet achieves the best on the above three different image features. Following \citep{chen2021learning}, we calculate the average of the ranking results of local- and global-level inputs as the final hybrid-level ranking results of ensemble patterns. 
 
\begin{table}[t]
\begin{center}
\fontsize{8}{9.5}\selectfont
\renewcommand\tabcolsep{5pt}
\caption{Comparisons of performances on Flickr30K 1K test set.  $*$ indicates the performance metrics attributed to the ensemble model. The best results are indicated in bold. For clearer comparison, our ensemble model is shown with a blue background and the improvement of the best contrasting method (with underline) is marked.} \label{tab:tab1_f30k}
\begin{tabular}{c|cccccc|c}
\hline

\hline
  \multicolumn{1}{c|}{\multirow{2}{*}{Method}} & \multicolumn{3}{c}{Image-to-Sentence} & \multicolumn{3}{c|}{Sentence-to-Image}      & \multirow{2}{*}{rSum} \\
   & Recall@1        & Recall@5       & Recall@10      & Recall@1  & Recall@5 & \multicolumn{1}{c|}{Recall@10} &                        \\ \hline
\multicolumn{8}{c}{With region-based image representation} \\ 
IMRAM$\rm ^*$ \citep{chen2020imram}    & 74.1 & 93.0 & 96.6 & 53.9 & 79.4 & 87.2 & 484.2  \\ 
VSE$\rm \infty$ \citep{chen2021learning}  &  81.7 &  95.4 &  97.6 &  61.4 &  85.9 &  91.5 &  513.5 \\ 
DIME$\rm ^*$  \citep{qu2021dynamic}    &  81.0  & 95.9  & 98.4  & 63.6  & 88.1  & 93.0  & 520.0\\
VSRN++$\rm ^*$  \citep{li2022image}  & 79.2 & 94.6 & 97.5 & 60.6 & 85.6 & 91.4 & 508.9 \\ 
NAAF$\rm ^*$  \citep{zhang2022negative}   & 81.9  & 96.1  & 98.3  & 61.0  & 85.3  & 90.6  & 513.2\\ 
\underline{AME$\rm ^*$  \citep{li2022action}}    &  81.9 &  95.9 &  {98.5} &   {64.6} &   {88.7} &   {93.2} &   {522.8}\\
CMSEI$\rm ^*$ \citep{ge2023cross}  &  {82.3} &  {96.4} & \textbf{98.6} & 64.1 & 87.3 & 92.6 & 521.3\\ 
 CHAN \citep{pan2023fine}   &  80.6 &  96.1 &  97.8 &  63.9 &  87.5 &  92.6 &  518.5 \\
 RCTRN* \citep{li2023reservoir}  &  78.4  &  95.4 &  96.8 &  60.4 &  84.9 &  93.7 &  509.6 \\
 KIDRR* \citep{xie2023unifying}  & 80.2 & 94.9 & 98.0 & 61.5 & 84.5 & 90.1 & 509.2 \\
 EKDM* \citep{yang2023external}  &  82.3 &  96.5 &  98.5 &  61.5 &  86.0 &  90.9 &  515.7  \\

\cline{1-8} 
 \multicolumn{1}{c|}{ \textbf{3SHNet}}      &
   {82.0} & 
   {96.2} & 
   {98.3} & 
   {64.8} & 
  {87.3} & 
   {92.8} & 
   {521.4} \\
 \multicolumn{1}{c|}{\cellcolor{blue!10}\textbf{3SHNet$^*$}} & \cellcolor{blue!10}\textbf{84.7{$_{\tiny\text{+2.8}}$}} & \cellcolor{blue!10}\textbf{96.8{$_{\tiny\text{+0.9}}$}} & 
\cellcolor{blue!10}{98.0{$_{\tiny\text{-0.5}}$}} & 
\cellcolor{blue!10}\textbf{66.1{$_{\tiny\text{+1.5}}$}} & \cellcolor{blue!10}\textbf{88.7{$_{\tiny\text{+0.0}}$}} & 
\cellcolor{blue!10}\textbf{93.4{$_{\tiny\text{+0.2}}$}} & \cellcolor{blue!10}\textbf{527.8{$_{\tiny\text{+5.0}}$}} \\

\hline
\multicolumn{8}{c}{With grid-based image representation} \\ 
SCO$\rm ^*$  \citep{huang2018learning}  & 55.5 & 82.0 & 89.3 & 41.1 & 70.5 & 80.1 & 418.5\\
 \underline{VSE$\rm \infty$   \citep{chen2021learning} }  &   {81.5} &  \textbf{97.1} &   {98.5} &   {63.7} &   {88.3} &   {93.2} &   {522.3} \\ 
\cline{1-8} 
 \multicolumn{1}{c|}{ \textbf{3SHNet}}     &
   {83.9} & 
  {96.7} & 
  {97.9} &   {65.1} & 
   {88.6} & 
   {93.3} & 
   {525.5}\\
 \multicolumn{1}{c|}{\cellcolor{blue!10}\textbf{3SHNet$^*$}}& \cellcolor{blue!10}\textbf{84.9{$_{\tiny\text{+3.4}}$}} & 
\cellcolor{blue!10}{97.0{$_{\tiny\text{-0.1}}$}} &  
\cellcolor{blue!10}\textbf{98.5{$_{\tiny\text{+0.0}}$}} & 
\cellcolor{blue!10}\textbf{67.2{$_{\tiny\text{+3.5}}$}} & \cellcolor{blue!10}\textbf{89.6{$_{\tiny\text{+1.3}}$}} & \cellcolor{blue!10}\textbf{94.0{$_{\tiny\text{+0.8}}$}} & \cellcolor{blue!10}\textbf{531.3{$_{\tiny\text{+9.0}}$}}\\ 

\hline
\multicolumn{8}{c}{With region- and grid-based image representation} \\ 
 CRGN  \citep{zhang2020deep} & 70.5 & 91.2 & 94.9 & 50.3 & 77.7 & 85.2 & 469.8  \\
 MLSL \citep{li2021multi} & 72.2 & 92.4 & 98.2 & 56.8 & 83.3 & 91.3 & 494.2\\
 VSE$\rm \infty$  \citep{chen2021learning}   &   {85.3} &  97.2 &  98.9 &  66.7 &   {89.9} &  94.0 &   {532.0}\\
CMCAN   \citep{zhang2022show}  & 79.5  & 95.6  & 97.6  & 60.9  & 84.3  & 89.9  & 507.8\\
    \underline{Imp.$\rm ^*$ \citep{wu2022improving} }  &  84.5 &   {97.3} &   {99.0} &   {66.8} &  89.7 &   {94.3} &  531.6 \\ 
  RAAN* \citep{wang2023rare}   & 77.1  & 93.6  & 97.3  & 56.0  & 82.4  & 89.1  & 495.5 \\
  HGAN \citep{guo2023hgan}   & 80.3  & 96.5  & 98.3  & 62.3  & 87.8  & 93.1  & 518.3  \\
  
\cline{1-8} 
 \multicolumn{1}{c|}{ \textbf{3SHNet}} &
   {86.1} & 
   {97.6} & 
  {98.8} &   {68.6} & 
   {90.1} & 
   {94.4} & 
   {535.6}\\
 \multicolumn{1}{c|}{\cellcolor{blue!10}\textbf{3SHNet$^*$}} & 
\cellcolor{blue!10}\textbf{87.1{$_{\tiny\text{+2.6}}$}} & \cellcolor{blue!10}\textbf{98.2{$_{\tiny\text{+0.9}}$}} &  
\cellcolor{blue!10}\textbf{99.2{$_{\tiny\text{+0.2}}$}} & 
\cellcolor{blue!10}\textbf{69.5{$_{\tiny\text{+2.7}}$}} & \cellcolor{blue!10}\textbf{91.0{$_{\tiny\text{+1.3}}$}} & \cellcolor{blue!10}\textbf{94.7{$_{\tiny\text{+0.4}}$}} & \cellcolor{blue!10}\textbf{539.7{$_{\tiny\text{+8.1}}$}}  \\ 
\hline

\hline
\end{tabular}
\end{center}
\vspace{-0.8em}
\end{table}

        \subsubsection{Quantitative comparison on MS-COCO.}    
        Tab. \ref{tab:tab1_coco_1K} and Tab. \ref{tab:tab1_coco_5K} present the quantitative results on two distinct MS-COCO test sets, 5-folds 1K and full 5K (the latter is a larger retrieval test set comprising 5000 images and 25000 sentences). Our 3SHNet significantly exceeds existing state-of-the-art methods on all recall metrics with different visual features. Specifically, for region-based features, compared to the best text-dependent method CHAN \citep{pan2023fine} on MS-COCO 1K test sets, our single-model 3SHNet achieves improvements of 1.7\% and 2.2\% on Recall@1 of image-to-sentence retrieval and sentence-to-image retrieval, respectively. The ensemble model 3SHNet* also gets improvements of 3.4\% on rSum compared to the text-dependent ensemble method EKDM* \citep{yang2023external}. On the larger 5K test set, both our 3SHNet and 3SHNet* achieve significant improvements compared to CHAN \citep{pan2023fine}and DCIN* \citep{li2023towards} with 454.9${(+11.0)}$ and 460.2${(+17.2)}$ on rSum, respectively. Notably, our single-model 3SHNet outperforms the best grid-based retrieval model VSE$\rm \infty$ \citep{chen2021learning} on all metrics, achieving the highest rSum scores of 538.9${(+7.3)}$ and 459.0${(+19.0)}$ on MS-COCO 1K (5-folds) and full 5K test sets, respectively. By combining multi-level visual features, our 3SHNet significantly boosts the retrieval performance compared to the state-of-the-art Imp.*. For example, it improves 2.1\% on Recall@1 of image-to-sentence retrieval and 3.4\% on Recall@1 of sentence-to-image retrieval on the 1K test set, and 4.4\% and 3.5\% on the larger 5K test set, respectively. We also provide comprehensive comparisons on different batch sizes in Section \ref{batchs} to fully demonstrate our superiority.  
         
        \subsubsection{Quantitative comparison on Flickr30K.}
        Tab.  \ref{tab:tab1_f30k} shows the quantitative results on a different dataset, the Flickr30K test set, where the proposed 3SHNet outperforms the state-of-the-art studies on three types of visual representations with the impressive gains of rSum.  
        Specifically, when using region-based image representation to align images and sentences, our method respectively improves the state-of-the-art AME* \citep{li2022action} by 2.8\%, 1.5\% in terms of Recall@1 on image-to-sentence and sentence-to-image retrieval directions, and by 5.0\% on rSum.
        When using grid-based image representation, our 3SHNet still markedly exceeds other models on all metrics, where it outperforms the second-best  VSE$\rm \infty$ \citep{chen2021learning} by 9.0\% in terms of rSum. 
        By combining local and global image representation, the retrieval performances are significantly improved and our 3SHNet significantly outperforms all competing methods, e.g., improving 8.1\% on rSum compared with the second-best Imp.* model \citep{wu2022improving}. 
        These observations on the Flickr30K benchmark serve as additional evidence of the robustness and superiority of our retrieval model.
        
\begin{table}[t]
\scriptsize
\begin{center}
\fontsize{8}{10.5}\selectfont
\renewcommand\tabcolsep{5pt}
\caption{Results on generalizability across datasets from MS-COCO to Flickr30k. $*$ indicates the performance of the ensemble model. Results marked with $^{\natural}$ indicate that they are derived from the released pre-trained model in their published works. } \label{tab:tab_transfor}
\begin{tabular}{c|cccccc|c}
\hline

\hline
\multicolumn{1}{c|}{\multirow{2}{*}{Method}} & \multicolumn{3}{c}{Image-to-Sentence} & \multicolumn{3}{c|}{Sentence-to-Image} &\multicolumn{1}{c}{\multirow{2}{*}{Rsum}}  \\
\multicolumn{1}{c|}{}                        & Recall@1   & Recall@5  & Recall@10   & Recall@1   & Recall@5 & \multicolumn{1}{c|}{Recall@10}  \\ \hline
    CVSE \citep{wang2020consensus}	   & 56.4 & 83.0 & 89.0	& 39.9	& 68.6 & 77.2 & 414.1\\
    SGR   \citep{qi2021self}    & 51.4 & 79.2 & 87.2 & 40.5 & 68.6 & 77.7 & 404.6\\ 
     SGRAF$^*$   \citep{diao2021similarity}  & 65.7 & 87.2 & 93.4 & 48.1 & 73.9 & 81.9 & 450.2 \\
     VSE$\rm \infty^{\natural}$  \citep{chen2021learning} &   {68.0} & 89.2	&  93.7	&  50.0	&  77.0 &  84.9 &  462.8 \\
    DIME$\rm ^{\natural}$  \citep{qu2021dynamic}    & 63.5 & 86.9 & 93.1 & 49.7 & 76.1 & 83.9 & 453.2\\
    DIME$\rm ^{*\natural}$  \citep{qu2021dynamic}   & 67.4	&  {90.1} &  {94.5}	&  {53.7}	&  {79.2} &  {86.5}  &  {471.4}\\
     ESA   \citep{zhu2023esa}    & 69.5 & 89.1 & 93.8 & 51.5 & 77.9 & 85.7 & 467.4\\ 
    \hline 	
    \textbf{3SHNet$^*$ (Region)} & \textbf{72.7} & \textbf{90.9} & {94.2} & \textbf{54.5} & \textbf{79.5} & \textbf{86.8} & \textbf{478.6} \\
     \textbf{3SHNet (Grid)}   &  \textbf{70.9} &  \textbf{91.6} &  \textbf{94.9} &  \textbf{53.7} &  \textbf{79.6} &  \textbf{87.0} &  \textbf{477.8} \\
    \textbf{3SHNet (Region+Grid)}    & \textbf{74.9} & \textbf{93.2} & \textbf{96.2} & \textbf{55.8} & \textbf{81.4} & \textbf{88.5}  & \textbf{490.0}
     \\ \hline  
     
\hline
\end{tabular}
\end{center}
\vspace{-2em}
\end{table}

\begin{figure}[t] 
	\centering
	\includegraphics[width=0.6\linewidth]{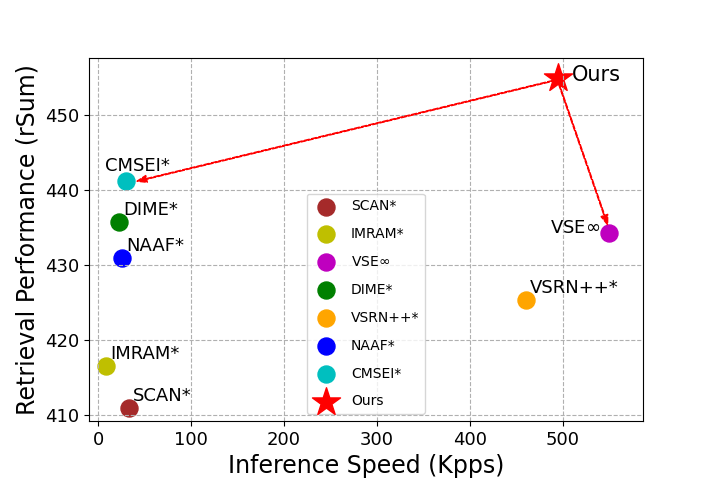}
	\caption{Inference speed (Kpps \citep{wang2022coder} means the number of image/sentence queries completed per second) and performance on MS-COCO 5K test set for image-text retrieval on single GPU (upper right is better). }
	\label{fig:Inf_speed}
\end{figure}

        \subsection{Generalization Capability for Cross-dataset Adaptation} 
        Generalization is one crucial and practical capability for cross-modal retrieval. Due to modality independence, 3SHNet is expected to acquire better generalization. To evaluate the generalization capability of our proposed visual semantic-spatial self-highlighting network, we create a cross-dataset transferring evaluation by pre-training methods on MS-COCO and validating them on Flickr30K test set in Tab. \ref{tab:tab_transfor}, which turns out to be more abundant than the existing experimental settings. 
        MS-COCO \citep{lin2014microsoft} and Flickr30K \citep{young2014f30k} have certain inherent differences in textual-annotation quality and inherent homogeneity of images, although they are both real-world datasets. As revealed by \citet{guan2018sequential,yu2023dataset}, Flickr30K sentence descriptions are more diverse while MS-COCO pays more attention to the consistency of image content.  
        Therefore, training on MS-COCO with testing on Flickr30K can reveal the zero-shot generalization capability of our proposed model when we change different textual-description domains.  
        Specifically, Tab. \ref{tab:tab_transfor} shows that 3SHNet alternatives accomplish the best. Especially among transferring models, single-model 3SHNet outperforms SGR   \citep{qi2021self}, CVSE \citep{wang2020consensus}, VSE$\rm \infty$ \citep{chen2021learning}, DIME \citep{qu2021dynamic} and  ESA \citep{zhu2023esa}, while ensemble-model 3SHNet* surpasses SGRAF*  \citep{diao2021similarity} and DIME* significantly. These performance gains reflect not only the conspicuous generalization ability of 3SHNet but also the superiority of our visual inner multi-modal interaction.

         \subsection{Inference Speed}
         
         To evaluate the efficiency of 3SHNet, we report both retrieval performances and off-line inference speeds in Fig. \ref{fig:Inf_speed}. 
         Following the existing methods \citep{lee2018stacked,chen2020imram,qu2021dynamic,zhang2022negative,ge2023cross}, we employ the caching strategy to exclude the non-real-time calculation of pre-stored features since reducing the repeated interactive calculation of images with texts to improve the real-time inference speed is one of our focuses. In our 3SHNet, the feature computing in our entire visual branch and textual branch can be taken as a feature pre-extraction process that benefited from modality independence. Thus, the ultimate inference speed will be found in the practical operation. Indeed, this is also an urgent demand in practical applications, such as multi-modal recommendation systems \citep{karedla1994caching,chen2020mobile}, since the feature pre-extraction can be processed offline. 
         Compared with the text-dependent methods, SCAN*, IMRAM*, DIME*, NAAF* and CMSEI* and modality-independent methods, VSRN++*, VSE$\rm \infty$, our 3SHRNet achieves comprehensive advantage on both performance and efficiency. For example, 3SHRNet is nearly 10 times faster than CMSEI* with the improvement of 15 points in performance. These superior results reflect the effective inner multi-modal interaction and modality guidance in 3SHRNet under the modality independence. 

\begin{table*}[t]
\begin{center}
\fontsize{8.0}{10.5}\selectfont
\renewcommand\tabcolsep{1.5pt}
\caption{Ablation studies on MS-COCO 1K (5-folds) and full-5K test sets. $R@$ is the abbreviation of $Recall@$. } \label{tab:tab3_ab}
\begin{tabular}{ccccc|cccccc|c|cccccc|c}
\hline

\hline
 \multicolumn{5}{c|}{\multirow{1}{*}{Method}}  &  \multicolumn{7}{c|}{MS-COCO 1K (5-folds)}  &  \multicolumn{7}{c}{MS-COCO 5K}\\  \hline
\multirow{2}{*}{NO.} & \multirow{2}{*}{Reg.} & \multirow{2}{*}{VSpM} & \multirow{2}{*}{VSeM} & \multirow{2}{*}{Seg.}   & \multicolumn{3}{c}{Image-to-Sentence} & \multicolumn{3}{c|}{Sentence-to-Image}      & \multirow{2}{*}{rSum} & \multicolumn{3}{c}{Image-to-Sentence} & \multicolumn{3}{c|}{Sentence-to-Image}      & \multirow{2}{*}{rSum} \\
 \multicolumn{5}{c|}{} & R@1 & R@5  & R@10 & R@1  & R@5 & R@10 &  & R@1 & R@5  & R@10  & R@1  & R@5 & R@10 &                      \\ \hline
1& \cellcolor{gray!10}$\surd$ & \cellcolor{gray!10}$\surd$ & \cellcolor{gray!10}$\surd$& \cellcolor{gray!10}$\surd$    & \cellcolor{gray!10}\textbf{83.1} & 
  \cellcolor{gray!10}\textbf{97.2} & 
 \cellcolor{gray!10}\textbf{99.3} & 
 \cellcolor{gray!10}\textbf{68.7} & 
 \cellcolor{gray!10}\textbf{92.4} & 
 \cellcolor{gray!10}\textbf{96.6} & 
 \cellcolor{gray!10}\textbf{537.3}  & 
 \cellcolor{gray!10}\textbf{63.8}  & 
 \cellcolor{gray!10}\textbf{88.1}  & 
 \cellcolor{gray!10}\textbf{94.0}  & 
 \cellcolor{gray!10}\textbf{47.0}  & 
 \cellcolor{gray!10}\textbf{76.6}  & 
 \cellcolor{gray!10}\textbf{85.4}  & 
 \cellcolor{gray!10}\textbf{454.9} \\ \hline
2&  $\surd$& $\surd$ & - & $\surd$       &  82.7 & 97.3 & 99.0 & 68.3 & 92.4 & 96.7 & 536.4 & 63.3 & 87.8 & 93.3  & 46.6 & 75.9 & 85.4 & 452.4\\
3&  $\surd$& - & $\surd$ & $\surd$   & 81.8 & 97.0 & 99.0 & 68.6 & 92.4 & 96.5 & 535.4 & 62.6 & 87.1 & 93.6 & 47.1 & 75.9 & 85.3 & 451.6\\ \hline
4&  $\surd$& $\surd$ & $\surd$ & -   & 82.1 & 97.1 & 99.2 & 67.8 & 92.3 & 96.5 & 535.0 & 61.7 & 87.3 & 93.7 & 46.1 & 75.8 & 85.0  & 449.6 \\ 
5& $\surd$&-& - & $\surd$   & 81.7 & 96.9 & 99.0 & 67.1 & 92.2 & 96.6 & 533.4  & 61.3 & 87.2 & 93.7 & 46.2 & 75.7 & 85.2 & 449.2\\ \hline
6& $\surd$& - & - & -   & 81.0 & 96.6 & 98.9  & 65.9 & 91.3 & 96.0 & 529.8 &  61.7 & 87.0 & 93.1   & 44.1 & 73.6 & 83.5 & 442.9\\ 
7& -& - & - & $\surd$  & 55.8 & 82.3 & 90.2 & 44.9 & 77.4 & 87.3 & 437.9 & 31.0 & 58.8 & 70.7 & 24.1 & 51.7 & 64.4 & 300.6\\ 
\hline

\hline
\end{tabular}
\end{center}
\end{table*}

    \subsection{Ablation Studies}
   
    \subsubsection{The effectiveness of semantic-spatial self-highlighting.} 
    We conduct different feature combinations to observe the performances of different feature combinations and evaluate the superior of our semantic-spatial self-highlighting method. 
    In Tab. \ref{tab:tab3_ab}, the comparison of No.6 and No.5  shows that segmentation features (Seg.) do contribute to the retrieval performance. But when simultaneously comparing No.7 and No.5, we find that the contribution of segmentation features is far below the one of the regular region-based local-level features. 
    Besides, by comparing No.4 and No.5 that fuse region and segmentation features respectively by feature concatenation and by our semantic-spatial self-highlighting, we find that the visual multimodal interactive features from VSpM and VSeM are superior to simple fusion features, \emph{e.g.} No.4 \emph{vs.} No.5 on rSum gets 535.0 \emph{vs.} 533.4  and 449.6 \emph{vs.} 449.2  on MS-COCO 1K and 5K test sets respectively. These reflect that: (1) segmentation features play an indecisive role in semantic richness, and (2) our semantic-spatial self-highlighting method does promote the effective embedding of segmentation features under the multimodal interaction. 
    \begin{figure}[t] 
    	\centering
            \vspace{-0.5em}  
    	\includegraphics[width=0.75\linewidth]{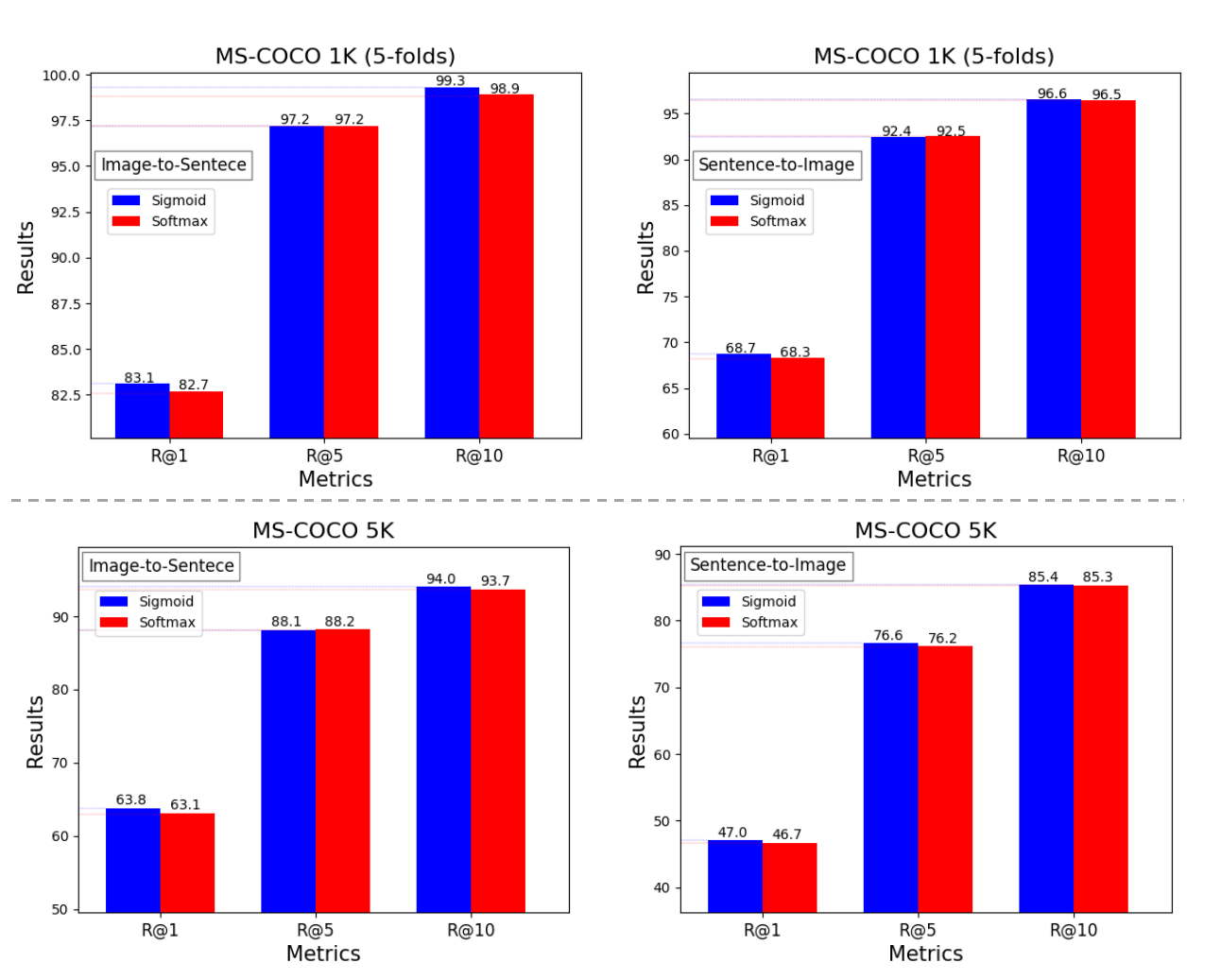}
    	\vspace{-0.5em}
    	\caption{ Comparisons of our proposed 3SHNet with different activate functions (Sigmoid \citep{mcculloch1943logical} \textit{VS.} Softmax \citep{chorowski2015attention}) in visual-semantic modelling. $R@$ is the abbreviation of $Recall@$.} 
    	\label{fig:sigmoid}
    \end{figure}   
    
    \subsubsection{Effects of visual-semantic modelling and visual-spatial modelling.} 
    To evaluate the impact of proposed visual-semantic modelling (VSeM) and visual-spatial modelling (VSpM) on image-sentence retrieval, we remove the visual-semantic salience embedding from VSeM and the visual-spatial embedding from VSpM, respectively.
    As shown in Tab. \ref{tab:tab3_ab}, the proposed approach makes absolute 1.9\% and 3.3\% drops in terms of rSum on MS-COCO 1K and 5K test sets when removing the visual-semantic salience embedding of visual-semantic modelling (indicated in NO. 3). 
    And it decreases absolutely 0.9\% and 2.5\% on rSum when removing the visual-spatial modelling (indicated in NO. 2). 
    These results manifest that our proposed visual semantic-spatial self-highlighting network can improve the distinguishability of image representations via the visual semantic and spatial interactions with corresponding segmentations. 

     \subsubsection{Effects of different activation functions.} \label{sigvssoft}
        Different activation functions have different meanings in Eq. (2) for visual-semantic modelling (VSeM). We chose Sigmoid \citep{mcculloch1943logical} because there may be several equally important objects in an image, which should all be attended to and should not be given less attention due to the weight limitation of Softmax \citep{chorowski2015attention}. 
        Additionally, the goal of the cosine similarity used in Sigmoid function is to enhance the differentiation among the marginal region features for more complete saliency. Since all the region proposals, including the marginal region proposals are related to the main semantic. It's unreasonable to take the center object region proposals as the same and with high semantic relation and take the marginal region proposals as the same and with little semantic relation. Indeed, to relieve the problem that the similarity value range is [-1,1] instead of (-$\infty$,+$\infty$), we follow a generic operation to divide similarity by $\sqrt{D}$ before the Sigmoid operation in practice. 
        Fig. \ref{fig:sigmoid} shows experimental results using different activation functions on the MS-COCO 1K and 5K test sets. 
        These observations demonstrate two aspects: (i) Both activation functions can motivate the effectiveness of the proposed visual-semantic modelling, which can obtain better results than the current state-of-the-art methods. (ii) As mentioned above, the Sigmoid function can activate objects with equal saliency as much as possible and enhance the differentiation among the marginal region features for more complete saliency via cosine similarity, allowing it to achieve slightly better experimental results than the Softmax function used in VSeM.
        
\begin{table*}[H]
\scriptsize
\begin{center}
\fontsize{8}{10.5}\selectfont
\caption{Results of the proposed 3SHNet on MSCOCO 5-fold 1K test set with different segmentation features extracted from different segmentation models, i.e., UPSNet \citep{xiong2019upsnet}  and PanopticFPN \citep{kirillov2019panoptic}.  $^{\natural}$ indicate that they are derived from the released pre-trained model in their published works.} \label{tab:tab_seg}
\begin{tabular}{ccc|cccccc|c}
\hline

\hline
\multicolumn{1}{c|}{\multirow{2}{*}{Method}} &\multicolumn{2}{c|}{\multirow{1}{*}{Segmentation}} & \multicolumn{3}{c}{Image-to-Sentence} & \multicolumn{3}{c|}{Sentence-to-Image} &\multicolumn{1}{c}{\multirow{2}{*}{Rsum}}  \\
\multicolumn{1}{c|}{}     &\multicolumn{1}{c}{Model}    &\multicolumn{1}{c|}{Performance}    & Recall@1   & Recall@5  & Recall@10   & Recall@1   & Recall@5 & \multicolumn{1}{c|}{Recall@10}  \\ \hline
\multicolumn{1}{c|}{\multirow{2}{*}{3SHNet}} & UPSNet  & 43.2$\rm ^{\natural}$     &   {83.1} &   {97.2} & 
   {99.3} &   {68.7} &  {92.4} &  96.6 &   {537.3}\\
\multicolumn{1}{c|}{}  & PanopticFPN  & 41.5$\rm ^{\natural}$    &  82.9 & 97.1 & 99.1 & 68.4 & 92.5 & 96.6 & 536.6\\
    \hline 	

\hline
\end{tabular}
\end{center}
\vspace{-1em}
\end{table*}

 \subsubsection{Effects of different segmentation models.} \label{segms}
 we execute the comparison experiments on two representative segmentation methods \citep{xiong2019upsnet, kirillov2019panoptic} to adequately evaluate the impact of the segmentation model's performance on the overall efficacy of the proposed model. As shown in Table \ref{tab:tab_seg}, when changing the segmentation model UPSNet \citep{xiong2019upsnet} to the lower-performing PanopticFPN \citep{kirillov2019panoptic}, the retrieval results decrease slightly from 537.3 to 536.6 in terms of rSum on MS-COCO test set, but is still ahead of existing state-of-the-art methods. 
        In addition, compared with the baseline model without segmentation information (No. 6 in Table \ref{tab:tab3_ab}), PanopticFPN-based 3SHNet also leads to a performance improvement of 6.8\% in terms of rSum score. These observations indicate that segmentation information from different segmentation models can effectively improve visual representation for image-sentence retrieval, and the changes of the segmentation model have relatively little impact on the final retrieval effect. The main reason is that the semantic image features play a leading role, while the segmentation information only plays a guiding role, \emph{i.e.} modeling the salient objects of the image from the semantic and spatial levels. Therefore, while ensuring a certain segmentation capability, the retrieval effect is not significantly affected, which relies less on segmentation performance.
      
    \begin{figure}[t] 
	\centering
	\includegraphics[width=1.0\linewidth]{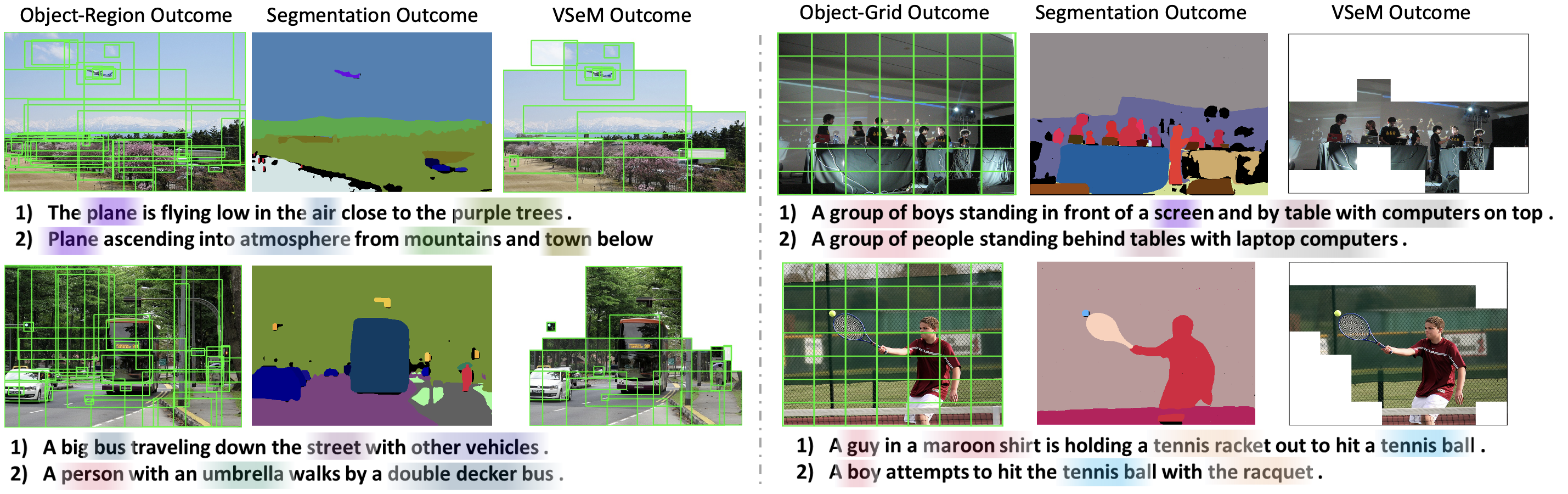}
	\caption{alient region-level (on the left) and grid-level (on the right) object visualizations from visual-semantic multimodal modelling (VSeM) guided by segmentations on MS-COCO dataset. Each visualization contains a visual image containing the original object outcome, its segmentation outcome and the corresponding VSeM outcome, and two random matching sentences with object highlights. The greater the salience of objects, the greater the transparency (best viewed in color).} %
	\label{fig:rgn_vse_attn}
    \end{figure}   
  
\begin{figure}[t] 
	\centering
	\includegraphics[width=0.9\linewidth]{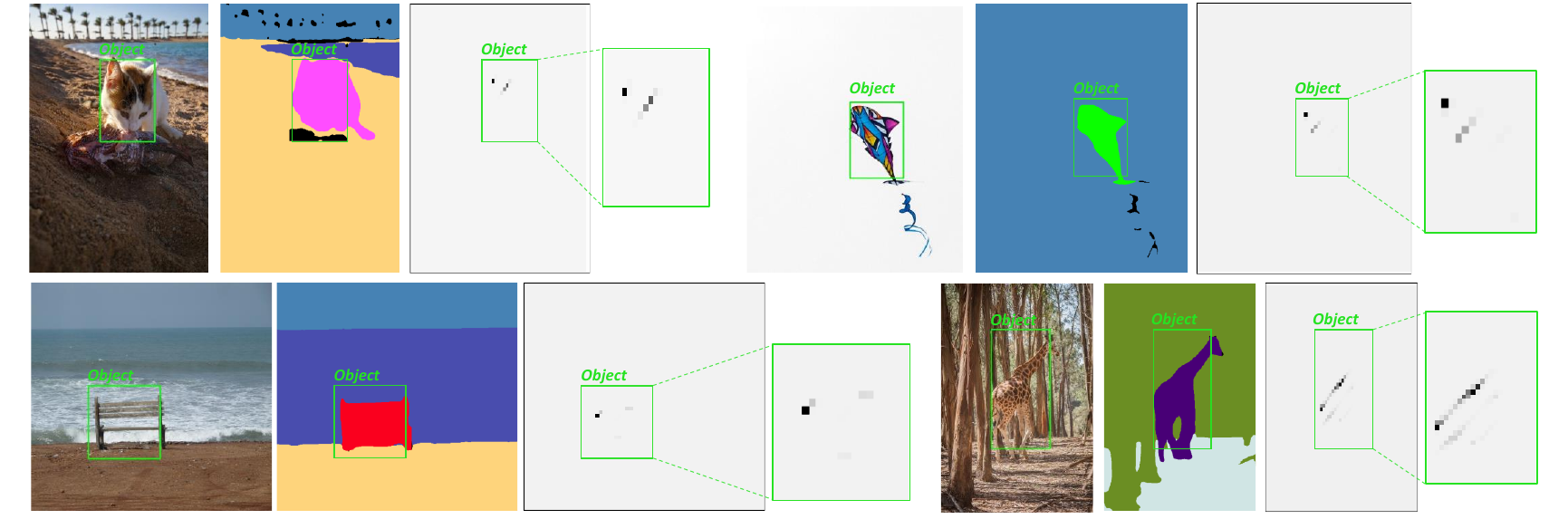}
	\caption{Visualization of salient spatial locations of corresponding salient objects on MS-COCO by visual-spatial multimodal modelling (VSpM). The greater the salience, the more pronounced the black colour (best viewed in color).}%
	\label{fig:vsp_attn}
\end{figure}  
\begin{figure}[!ht] 
	\centering
	\vspace{1em}
	\includegraphics[width=1\linewidth]{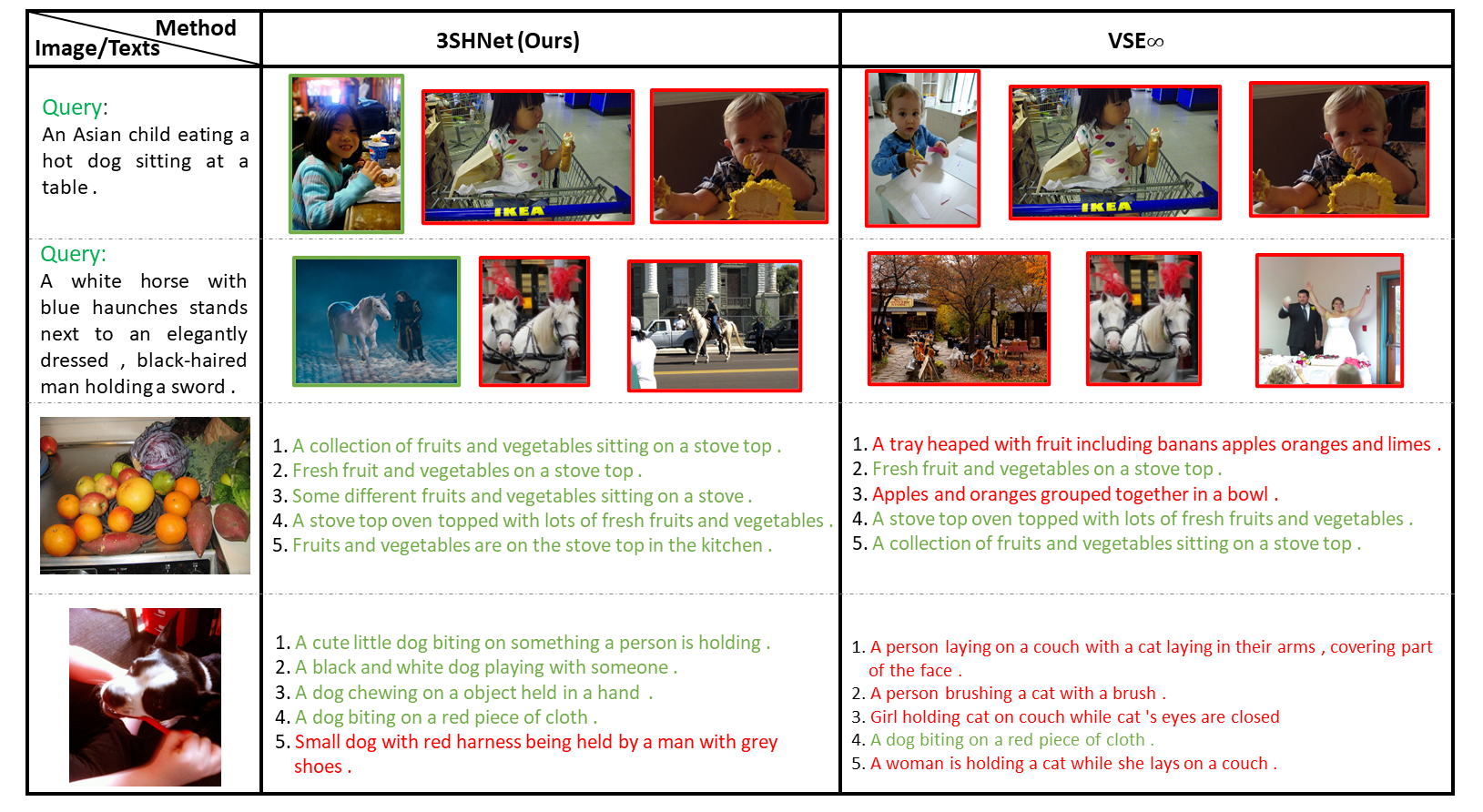}
	\caption{Comparisons of sentence-to-image and image-to-sentence retrieval between our 3SHNet and VSE$\infty$ \citep{chen2021learning} on MS-COCO test set. For image retrieval, we showcase the foremost three ranked images, ordered in a left-to-right ranking fashion. Ranked images that align correctly are denoted in green, while any discordant matches are denoted in red.  For image-to-sentence retrieval, we display the top five sentences retrieved for each image query, with any mismatches distinctly accentuated in red (best viewed in color).} 
	\label{fig:example_t2i_supply}
\end{figure}
\begin{figure}[!hb] 
    	\centering
    	\includegraphics[width=1\linewidth]{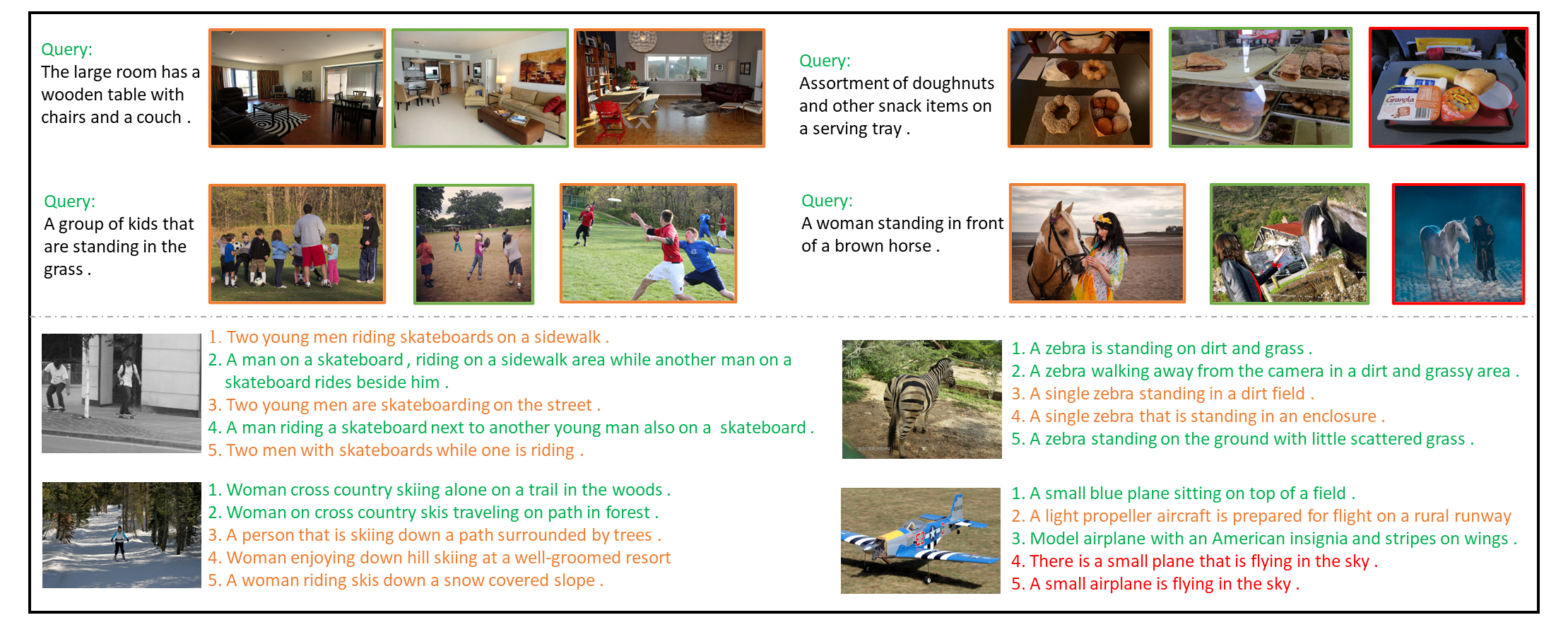}
    	\caption{Visualization of the unsuccessful image-retrieval examples (at the top) and sentence-retrieval examples (at the bottom) on MS-COCO by the proposed 3SHNet. In these cases, some mismatched images/sentences (in orange) in datasets have similar semantic content to true matches (in green), which can also express the main content of the corresponding sentence/image, even more appropriately than a correct match result. Even completely incorrect retrievals (in red) are ranked lower than the correct results.}
    	\label{fig:badcase_supply}
    	\vspace{1em}
    \end{figure}      

        \subsection{Qualitative Analysis} 
        To further understand the contribution of visual semantic and spatial salience embedding from our proposed VSeM and VSpM, we visualize the salient regions and girds used to represent the main content of images in Fig. \ref{fig:rgn_vse_attn} and visualize the spatial embedding weights of the most salient corresponding regions in Fig. \ref{fig:vsp_attn}. 
        It is clear from Fig. \ref{fig:rgn_vse_attn} that our visual-semantic modelling can concentrate on the main regions and grids containing salient objects guided by corresponding segmentations to improve image representation capabilities. 
        In addition, we also visualize some examples in Fig. \ref{fig:vsp_attn} to help understand the effectiveness and interpretability of our visual-spatial modelling. 
        Accurate visual-spatial embeddings can further enhance the ability and distinguishability of salient object representations in images. 
        
        Furthermore, more visualizations and analyses of two-modality retrieval cases are available for comprehensive comparisons in Fig. \ref{fig:example_t2i_supply}.  For image retrieval, we exhibit the top three ranked images corresponding to each sentence query by our proposed 3SHNet and the latest comparison VSE$\infty$ \citep{chen2021learning}, respectively. True matches are highlighted within green-bordered boxes, while incorrect matches are indicated by red borders. 
        In addition, we also show the image-to-sentence retrieval results (top-3 retrieved sentences) forecasted by our 3SHNet and VSE$\infty$ \citep{chen2021learning}, where instances of discrepancy are highlighted in red. 
        These observations demonstrate that our approach can obtain more accurate search results. 
        Moreover, we show some bad cases of image and sentence retrieval tasks in Fig. \ref{fig:badcase_supply}. 
        In these cases, most of the wrong results (in orange) have similar content to the true matches (in green), which should also be valid retrievals since they express the main content of the corresponding sentence/image even more accurately than the correct matches. 
       \begin{figure}[t] 
    	\centering
    	\includegraphics[width=1\linewidth]{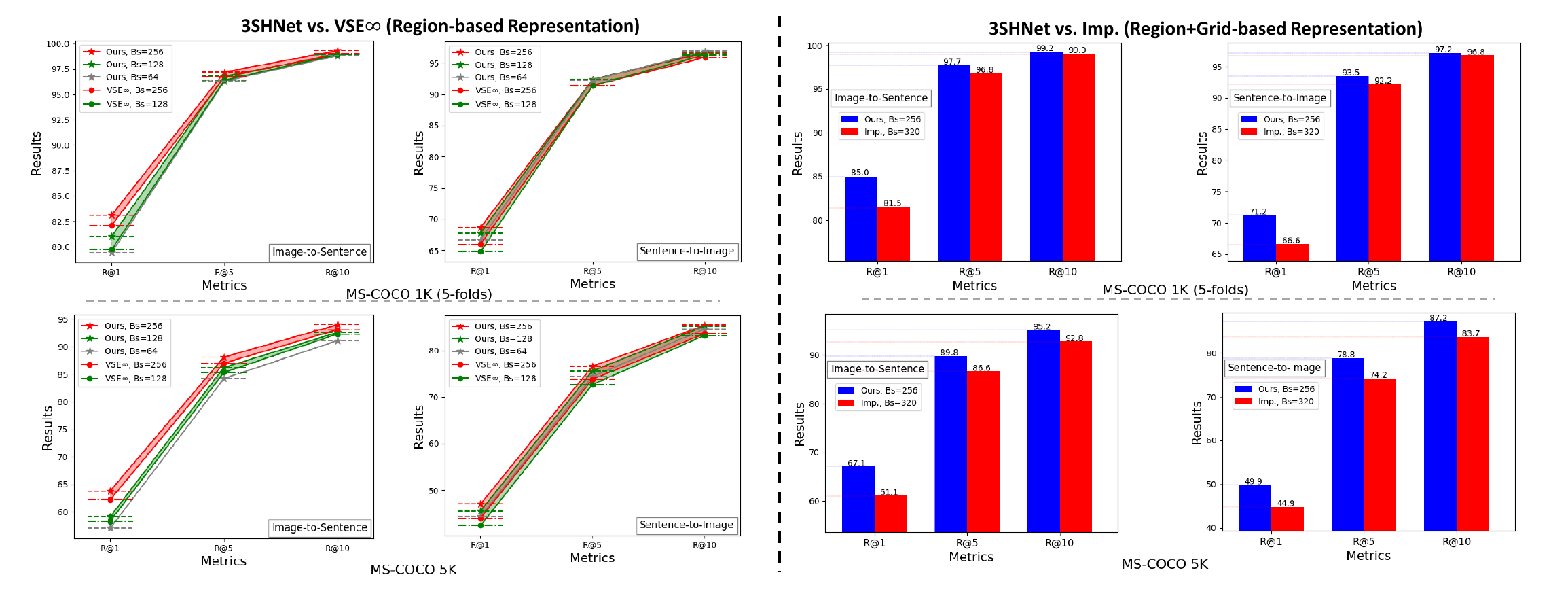}
    	\vspace{-1.5em}
    	\caption{Comparisons of our proposed 3SHNet and two state-of-the-art methods (VSE$\infty$ \citep{chen2021learning} and Imp. \citep{wu2022improving}) in different batch sizes used the same visual representations, respectively. $R@$ is short for $Recall@$.} 
    	\label{fig:batchsize}
    \end{figure}  
    \begin{table}[t]
    \scriptsize
    \begin{center}
    \fontsize{8}{11}\selectfont
    \renewcommand\tabcolsep{5pt}
    \caption{Comparisons between the proposed 3SHNet and some large-scale pre-trained visual-language methods on MS-COCO full-5K test set. $z$ means zero-shot cross-modal retrieval results, where the model is pre-trained on the large-scale image-sentence pairs. $\rm Bs$ means the mini-batch size. } \label{tab:vl}
    \begin{tabular}{c|c|c|c|cccc}
    \hline
    
    \hline
    \multicolumn{1}{c|}{\multirow{2}{*}{Method}} &\multicolumn{1}{c|}{\multirow{2}{*}{Pretrain}} &\multicolumn{1}{c|}{\multirow{2}{*}{GPUs}} &\multicolumn{1}{c|}{\multirow{2}{*}{Bs}} & \multicolumn{2}{c}{Sentence Retrieval} & \multicolumn{2}{c}{Image Retrieval}   \\
    \multicolumn{1}{c|}{}                  &    &    &   & Recall@1   & Recall@5    & Recall@1   & Recall@5  \\ \hline
        ViLBERT \citep{zhang2020learning}  & 3.3M &   8 TitanX  & 512 & 57.5 & 84.0 & 41.8 & 71.5 \\
        UNITER \citep{chen2020uniter}     & 9.6M & 16 V100 & 192  & 63.3 & 87.0 & 48.4 & 76.7  \\
        Unicoder \citep{li2020unicoder}  & 3.8M &  4 V100  & - & 62.3 & 87.1 & 46.7 & 76.0 \\ 
        OSCAR  \citep{li2020oscar}    & 6.5M & 16 V100 & 1,024 & 70.0 & 91.1 & 54.0 & 80.8  \\ 
        CLIP$_{z}$ \citep{radford2021learning} &  400M & 592 V100 & 32,768 & 58.4 & 81.5 & 37.8 & 62.4\\ 
        ALIGN$_{z}$ \citep{jia2021scaling}& 1.8B & 1024 TPUv3 & 16384 & 58.6 & 83.0 & 45.6 & 69.8\\ 
        ALIGN  \citep{jia2021scaling} & 1.8B & 1024 TPUv3 & 16384 & 77.0 & 93.5 &  59.9 & 83.3 \\ 
        ALBEF \citep{li2021align}  & 4.0M   & 8 A100   & 512   & 73.1 & 91.4  & 56.8 & 81.5 \\ \hline
        \textbf{3SHNet}  & N/A  &  1 TitanX  & 256 & {67.9} & {90.5}  & {50.3} & {79.3}
         \\ \hline  
         
    \hline
    \end{tabular}
    \end{center}
    \end{table}      
    \subsection{Discussion} 
    \subsubsection{Discussion on different batch sizes.} \label{batchs}
    The ISR literature shows that a larger batch size may improve retrieval performance. The main reason is that during the training process, the two-way triple ranking loss of the hard negative mining strategy\citep{faghri2017vse++} is used, so a larger batch can have a greater probability of containing high-quality negative samples, which better optimizes our objective function. 
    Since our model is free from the dependence on textual guidance, we can use a relatively high throughput under the same computing facilities. 
    As shown in Fig. \ref{fig:batchsize}, we report more results from our proposed 3SHNet on MS-COCO benchmark compared with the latest methods, \textit{i.e.,} VSE$\infty$ \citep{chen2021learning} and Imp. \citep{wu2022improving} used the same visual representations in different batch sizes. 
    These observations suggest that a larger mini-batch can somewhat improve the performance of cross-modal retrieval. 
    In addition, in contrast to the state-of-the-art methodologies in the same batch size, our 3SHNet outperforms them by a large margin in Fig. \ref{fig:batchsize} on all six recall metrics. Our proposed visual semantic-spatial self-highlighting network can boost the efficacy of image-sentence retrieval.
       
    \subsubsection{Discussion on large-scale pre-trained models.} \label{discuss}  
        Pre-trained visual language representations on large-scale datasets are becoming increasingly popular, especially in companies with large-scale parallel computing power. 
        However, due to the limitation of computation facility requirements, it is difficult to carry out large-scale pre-training in universities or research institutions. 
        For example, the pre-training of UNITER-base and UNITER-large in \citep{li2020unicoder} involved the utilization of 882 and 3645 V100 GPU hours, respectively. 
        Additionally, most of the excellent large-scale pre-training methods \citep{chen2020uniter,radford2021learning,jia2021scaling} also depend on extensive cross-modal interactions within vast collections of image-text pairs. 
        The text-dependent visual representation learning approach leads to a long inference retrieval time, which is hardly applied to real-life scenarios. 
        In this section, we report the results of our proposed approach compared to some popular methods, such as ViLBERT \citep{zhang2020learning}, UNITER \citep{chen2020uniter}, Unicoder \citep{li2020unicoder}, OSCAR \citep{li2020oscar}, CLIP \citep{radford2021learning}, ALIGN \citep{jia2021scaling} and ALBEF \citep{li2021align} that pre-trained on large-scale datasets. 
        As shown in Table \ref{tab:vl}, compared to the large-scale visual-language pre-trained methods, our approach achieves competitive results at a smaller computation facility requirement without large-scale visual-language pre-training. 
        For example, the performances of our 3SHNet are better than UNITER \citep{chen2020uniter}, requiring 16 V100 GPUs in six evaluation metrics. 
        In addition, as mentioned in \citet{chen2021learning}, VSE \citep{faghri2017vse++,chen2021learning} methods (the same framework as our 3SHNet) exhibit significantly enhanced speed in large-scale multi-modal retrieval due to the expeditious pre-computed computation or indexing of holistic embeddings \citep{johnson2019billion}.

        Although our approach is relatively lower than ALIGN \citep{jia2021scaling}, which has stronger computation facilities and a larger number of image-text pairs at larger batch sizes, we propose a scheme that can apply our proposed method on large-scale datasets, which we will explore further in future work. 
        Specifically, recently, the introduction of \citet{kirillov2023segment} makes it easy and fast to obtain semantic segmentation results in arbitrary scenarios, which will facilitate the application of our proposed VSeM and VSpM to large-scale datasets. 

    \section{Theoretical and Practical Implications}
    
    We propose 3SHNet for image-sentence retrieval that introduces a novel segmentation-based visual semantic-spatial self-highlighting schema into an end-to-end modality-independence modelling cross-modal alignment framework. The mainly visual-semantic and visual-spatial multimodal interactions mine the semantic saliency and spatial saliency of visual objects respectively, thereby improving the discriminability of image representations during the cross-modality alignment process. Guidance information from semantic segmentation overcomes the lack of textual dependence and maintains modality independence, thereby ensuring retrieval efficiency. 

    For theoretical implications, the proposed 3SHNet overcomes the obvious shortcomings of the two existing mainstream methods, i.e., low efficiency and low generalization due to the deep textual dependence in textual-guidance-based visual representation learning methods \citep{chen2020imram,qu2021dynamic,ge2023cross,pan2023fine} and the ignoring human-like attention on the prominent objects and their locations in the visual hybrid-level representation enhancing methods \citep{liu2018dense,ge2021structured,chen2021learning,ma2023beat}.
    In particular, our 3SHNet introduces the segmentation information to highlight the semantic saliency and spatial saliency of objects within the visual modality, which can replace the complex visual-textual interaction operations to keep the retrieval high-efficiency and improve the salience of prominent objects and their locations in the visual hybrid-level representations to improve the retrieval performance. 

    For practical implications, our 3SHNet aims to construct a high-precision, high-efficiency, and high-generalisation image-sentence retrieval model.
    This ensures retrieval efficiency while ensuring retrieval performance, thus providing the possibility for practical applications, such as multi-modal retrieval within search engines \citep{he2011using}. It can be applied to both large websites and private systems, such as library multimedia systems \citep{lee2003multimedia}.  
    Furthermore, 3SHNet does not rely on large-scale computing resources, thus ensuring its portability to new private data.

\section{Conclusion} 
In this paper, we enhance visual representation under the modality-independent pattern for high-precision, high-efficiency, and high-generalization image-sentence retrieval, where the visual semantic salience and spatial locations are highlighted based on visual segmentations. 
Specially, the visual-semantic and visual-spatial multimodal interactions are designed in our 3SHNet based on a two-tower modality-independent framework, which involves the hybrid-level visual representations, \textit{i.e.,} local-level region-base feature and global-level grid-based feature.
The superiority of our 3SHNet is evident in extensive quantitative comparisons, showcasing its state-of-the-art performance and efficiency across popular benchmarks such as MS-COCO and Flickr30K under various evaluation metrics.

\section{Acknowledgements}
Fuhai Chen's research was supported in part by the Fujian Provincial Department of Education Youth Project (grant JZ230006) and the Engineering Research Center of Big Data Intelligence, Ministry of Education, and Fujian Key Laboratory of Network Computing and Intelligent Information Processing (Fuzhou University); Xuri Ge’s research was supported in part by China Scholarship Council (CSC) from the Ministry of Education of China (No. 202006310028).


\printcredits

\bibliographystyle{apacite}
\bibliography{cas-refs}





\end{document}